\documentclass{article}

\usepackage{multirow}  
\usepackage{booktabs}   
\usepackage{graphicx}   
\usepackage{subcaption} 
\usepackage{tabularx}   
\usepackage{colortbl}   
\usepackage{rotating}   
\usepackage{bm}
\usepackage{dsfont}
\usepackage{algorithm}  
\usepackage{algorithmic}
\usepackage{wrapfig}
\usepackage{tikz}
\usetikzlibrary{shadings}
\usepackage[preprint]{neurips_2025}

\usepackage{enumitem}

\usepackage[utf8]{inputenc} 
\usepackage[T1]{fontenc}    
\usepackage{hyperref}       
\usepackage{url}            
\usepackage{booktabs}       
\usepackage{amsfonts}       
\usepackage{nicefrac}       
\usepackage{microtype}      
\usepackage{xcolor}         

\usepackage{amsmath} 
\newtheorem{proposition}{Proposition}[section]

\definecolor{new_blue}{RGB}{31,78,120}
\definecolor{new_red}{RGB}{136,14,37}

\title{Dynamic Graph Condensation}

\author{%
  Dong Chen$^{1, 2}$,
  \quad Shuai Zheng$^{1, 2}$,
  \quad Yeyu Yan$^{3, 4}$
  \quad Muhao Xu$^{1, 2}$,\\ 
  \quad \textbf{Zhenfeng Zhu}$^{1, 2}$\thanks{Corresponding author.}
  \quad \textbf{Yao Zhao}$^{1, 2}$
  \quad \textbf{Kunlun He}$^{3}$\\
  \\
  $^1$~Institute of Information Science, Beijing Jiaotong University \\
  $^2$~Visual Intelligence + X International Joint Laboratory of the Ministry of Education\\
  $^3$~Medical Big Data Research Center, Chinese PLA General Hospital Beijing, China \\
  \texttt{dchen2001@bjtu.edu.cn} \\
}

\begin{document}

\maketitle

\begin{abstract}

Recent research on deep graph learning has shifted from static to dynamic graphs, motivated by the evolving behaviors observed in complex real-world systems. However, the temporal extension in dynamic graphs poses significant data efficiency challenges, including increased data volume, high spatiotemporal redundancy, and reliance on costly dynamic graph neural networks (DGNNs). To alleviate the concerns, we pioneer the study of \textit{dynamic graph condensation (DGC)}, which aims to substantially reduce the scale of dynamic graphs for data-efficient DGNN training. Accordingly, we propose DyGC, a novel framework that condenses the real dynamic graph into a compact version while faithfully preserving the inherent spatiotemporal characteristics. Specifically, to endow synthetic graphs with realistic evolving structures, a novel spiking structure generation mechanism is introduced. It draws on the dynamic behavior of spiking neurons to model temporally-aware connectivity in dynamic graphs. Given the tightly coupled spatiotemporal dependencies, DyGC proposes a tailored distribution matching approach that first constructs a semantically rich state evolving field for dynamic graphs, and then performs fine-grained spatiotemporal state alignment to guide the optimization of the condensed graph. Experiments across multiple dynamic graph datasets and representative DGNN architectures demonstrate the effectiveness of DyGC. Notably, our method retains up to 96.2\% DGNN performance with only 0.5\% of the original graph size, and achieves up to 1846$\times$ training speedup.
\end{abstract}

\section{Introduction}

Graphs~\cite{graph, graph0deep} have been applied in a wide range of fields, including social network analysis, recommender systems and epidemiology. However,  graph in many industrial scenarios  are often large-scale, posing significant challenges for efficient data storage and the practical applications of graph neural networks (GNNs)~\cite{GNNSurvey, GCN, GraphSAGE}. To address this scalability issue, an advanced data-centric solution is graph condensation (GC)~\cite{GRsurvey, GCsurvey}. GC has attracted considerable attention in graph reduction due to its remarkable compression ratio and lossless performance. It aims to synthesize a compact yet informative graph that represents essential knowledge of the large real graph. The condensed graph enables efficient GNN training while maintaining performance comparable to full-graph training.

While significant progress has been made in static graph research, many real-world networks exhibit dynamic behaviors with valuable temporal information, necessitating the study of dynamic graphs~\cite{dgsurvey, dgsurvey2} where nodes and structures evolve over time. However, the introduction of the time dimension further exacerbates graph efficiency~\cite{dgeffi, SpikeNet} issues. The temporal extension of dynamic graphs increases data volume and necessitates computationally expensive dynamic GNNs (DGNNs)\cite{DGNNsurvey} to capture complex spatiotemporal dependencies. Moreover, dynamic graphs often exhibit pronounced spatiotemporal  redundancy\cite{temporalreb}, owing to the typically gradual and localized evolution of real-world systems.  These inefficiencies stem from the data itself, which highlights the necessity of data-centric solutions that improve the efficiency of dynamic graph learning.

\begin{wrapfigure}{r}{0.51\linewidth}
    \centering
    \vspace{-0mm}
    \includegraphics[width=\linewidth]{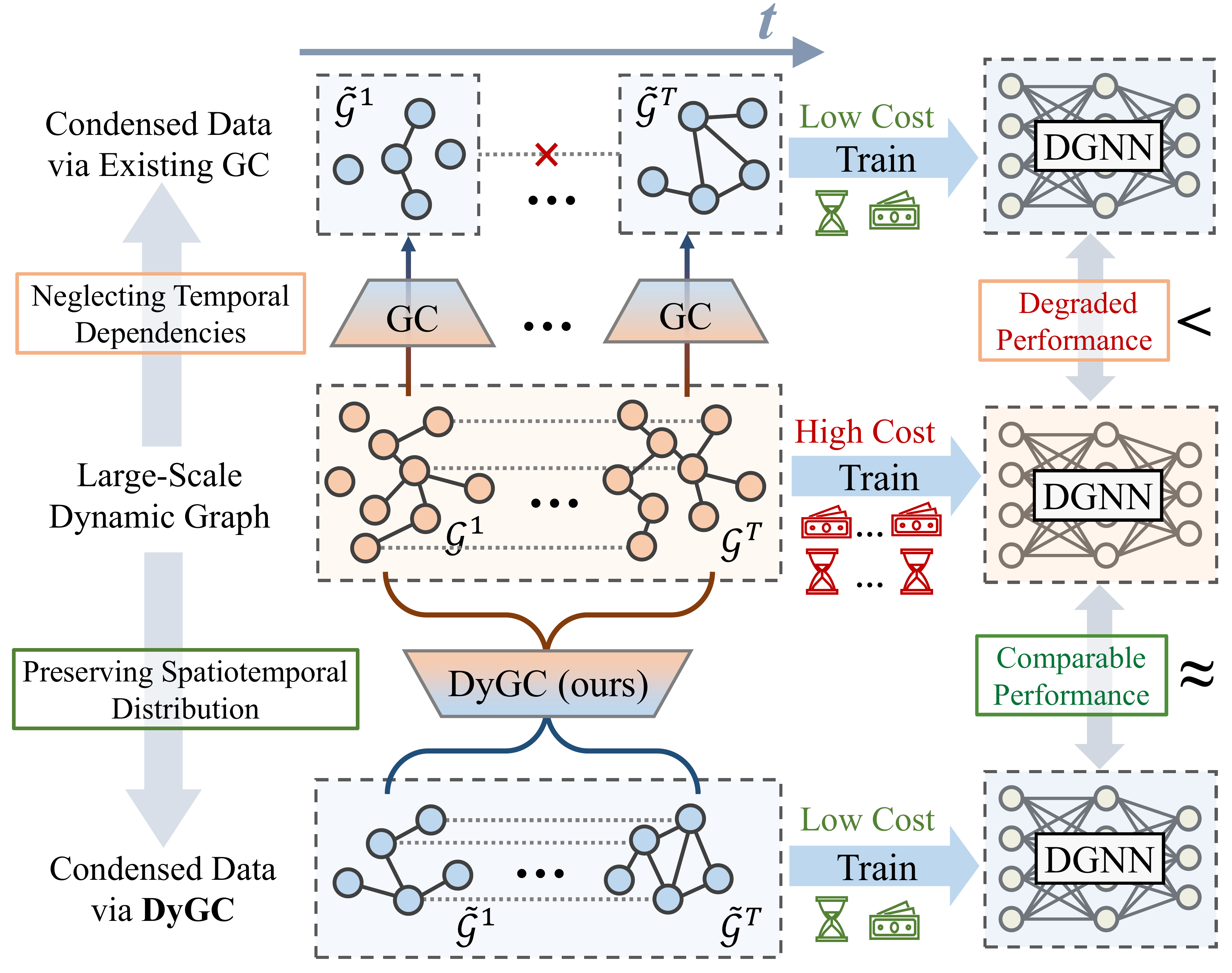}
    \caption{Comparisons of existing GC vs dynamic graph condensation (DyGC) on dynamic graphs.}  
    \label{intro}
\end{wrapfigure}

Unfortunately, although existing GC methods significantly improve data efficiency on static graphs, they show poor adaptability in dynamic settings. As illustrated in Figure~\ref{intro}, existing graph condensation methods can only perform snap-wise condensation when processing dynamic graphs. This approach not only fails to preserve temporal dependencies across snapshots, resulting in degraded DGNN performance, but also incurs significant computational overhead due to repeated condensation.
In practice, it is non-trivial to design an elegant method of condensation for dynamic graph. Unlike in the static setting, dynamic graphs exhibit intricate and tightly coupled spatiotemporal dependencies, where both node features and structural patterns evolve over time.  This necessitates tailored strategies to ensure that the condensed dynamic graph faithfully preserves the original spatiotemporal characteristics. Moreover, to realistically capture the structural evolution of dynamic graphs, structure generation in dynamic graph condensation should incorporate temporal modeling to prevent structural isolation across graph snapshots.


In light of these challenges, we explore the study of \textit{dynamic graph condensation (DGC)} and propose \textbf{DyGC}, the first framework designed to condense a compact dynamic graph for efficient DGNN training. Unlike existing GC methods, DyGC jointly incorporates structural and temporal perspectives to approximate the real dynamic graph by preserving its overall spatiotemporal distribution. Specifically, to model evolving structures in dynamic graphs, a spiking structure generation mechanism inspired by the dynamic behavior of spiking neurons in spiking neural networks (SNNs)~\cite{SNN} is introduced. Unlike conventional neurons, spiking neurons have temporal-aware neural activity~\cite{SNNtemporal} and binary event-driven outputs, enabling them to naturally model evolving connectivity patterns in dynamic graphs. To achieve high-fidelity preservation of spatiotemporal characteristics, DyGC introduces a tailored distribution matching approach for dynamic graphs. It first constructs a semantically rich state-evolving field to capture the dynamic graph's evolving spatiotemporal states, and then performs fine-grained state alignment to ensure the condensed graph faithfully preserves the real distribution.  In summary, the contributions of this work are listed as follows:
\begin{itemize}[itemsep=2pt,leftmargin=*,topsep=2pt]
    \item We make the first attempt to the study of dynamic graph condensation and propose an innovative framework named DyGC, which condenses large-scale dynamic graphs with intricate spatiotemporal dependencies into  small yet informative synthetic graphs.
    \item To endow synthetic graphs with evolving structures, we design a novel Spiking Structure Generation mechanism. It models the changing inter-node affinity in dynamic graphs as varying membrane voltage in spiking neurons and leverages the binary spiking behavior to induce connectivity.
    \item By aligning the diverse spatiotemporal states of dynamic graphs in a novel and semantically rich state evolving field, we achieve precise dynamic graph distribution matching that ensures the condensed graph faithfully preserve the inherent spatiotemporal characteristics.
    \item Extensive experiments across multiple  datasets and various DGNN architectures demonstrate the superior performance and generalization ability of DyGC.
\end{itemize}

\textbf{Prior Work.} Motivated by dataset distillation~\cite{dd}, many recent studies~\cite{GCond, SFGC, SGDD, SimGC, CGC} have explored the application of distillation techniques to graph-structured data, commonly referred to as graph condensation (GC). However, existing GC methods are primarily designed for static graphs and exhibit limited effectiveness when deal with dynamic graphs, as they fail to capture the crucial temporal dependencies. 
On the other hand, while dynamic graph neural networks (DGNNs)\cite{DGNNsurvey, T-GCN, DySAT, STGCN, EvolveGCN} have shown the effectiveness in processing dynamic graphs, these models often suffer from significant computational overhead due to the additional temporal components~\cite{LSTM, attention}, especially on the large-scale graphs.
This gap underscores the need for data condensation methods specifically tailored to dynamic graphs. A more detailed discussion of related work is provided in Appendix~\ref{apprw}.

\section{Preliminary} 
\textbf{Discrete-time Dynamic Graph.} We consider the task of dynamic graph condensation under a discrete-time setting, where a dynamic graph is represented as a sequence $\mathcal{T}=\{G^t \}_{t=1}^T \in \mathcal{G}$ with $T$ graph snapshots. The snapshot $G^t = (\mathcal{V}^t, \mathcal{E}^t)$ at time $t$ consists of a node set $\mathcal{V}^t = \{v_1, \dots, v_n\}$ and an edge set $\mathcal{E}^t \subseteq \mathcal{V}^t \times \mathcal{V}^t$.  The topology structure of snapshot $\mathcal{G}^t$ can be described by an adjacency matrix $\mathbf{A}^{(t)} \in \mathbb{R}^{n \times n}$ with $\mathbf{A}^{(t)}_{i,j}=1$ if $(i,j)\in \mathcal{E}^t$ or 0 otherwise. In each snapshot, nodes are paired with features $\mathbf{X}^{(t)} = \{x^{(t)}_v \mid \forall v \in \mathcal{V}^t\}\in \mathbb{R}^{n \times d}$. Without loss of generality, we assume that all snapshots share a consistent and fixed node set $\mathcal{V}_{\mathcal{T}}$, with nonexistent nodes treated as dangling ones with zero degree. Consequently, the evolving adjacency matrices and node features are denoted as $\mathcal{A} = [\mathbf{A}^{(t)}]_{t=1}^T \in \mathbb{R}^{T \times n \times n}$ and $\mathcal{X} = [\mathbf{X}^{(t)}]_{t=1}^T \in \mathbb{R}^{T \times n \times d}$, respectively. For the node set $\mathcal{V}_{\mathcal{T}}$, the corresponding label set is $\mathcal{Y} = \{y_1, \dots, y_n\} \in \{\emptyset, 1, \dots, C\}^n$ over $C$ classes, where $\emptyset$ indicates that the node is unlabeled. Thus, the dynamic graph is fully characterized by a triplet $\mathcal{T}=(\mathcal{A}, \mathcal{X}, \mathcal{Y} )$.

\textbf{Problem Formulation.} Given a large-scale dynamic graph $\mathcal{T}$, dynamic graph condensation (DGC) aims to synthesize a compact graph $\mathcal{S} = (\tilde{\mathcal{A}}, \tilde{\mathcal{X}}, \tilde{\mathcal{Y}} ) \in \mathcal{G}$, where $\tilde{\mathcal{A}} \in \mathbb{R}^{T \times m \times m}$, $\tilde{\mathcal{X}} \in \mathbb{R}^{T \times m \times d}$, $\tilde{\mathcal{Y}} = \{\tilde{y}_1, \dots, \tilde{y}_m\} \in \{1, \dots, C\}^m$, and $m \ll n$, such that a DGNN trained on $\mathcal{S}$ can achieve comparable performance to one trained on the much larger $\mathcal{T}$. In this work, we adopt the distribution matching paradigm to achieve dynamic graph condensation. Specifically, the condensed $\mathcal{S}$ is obtained by minimizing the distributional discrepancy between the synthetic and real dynamic graphs. This objective can be formulated as follows:
\begin{equation}
\mathcal{S}^* = \arg\min_\mathcal{S} \mathcal{D}(P_{\mathcal{T}}, P_{\mathcal{S}}),
\label{opt}
\end{equation} 
where $\mathcal{D}$ denotes a metric that quantifies the discrepancy between the distributions $P_{\mathcal{T}}$ and $P_{\mathcal{S}}$. In practice, a key challenge lies in meaningfully and accurately measuring the distributional discrepancy between two dynamic graphs, as it directly guides the optimization of the condensed graph.


\textbf{Leaky Integrate-and-Fire (LIF) Model.} LIF~\cite{LIF_2} is a typical neuron model in Spiking Neural Networks (SNNs)~\cite{LIF}. It mimics neuronal dynamics by accumulating membrane voltage like a charging capacitor. When the voltage exceeds a threshold, the neuron fires a spike signal and resets to a constant value $U_{\text{reset}}$. Additionally, this model includes a leak current term to simulate how the neuron's voltage slowly returns to the baseline level. The membrane voltage dynamics of a LIF neuron model can be described by the following differential equation:
\begin{equation}\label{LIF}
\tau_m \frac{dU}{dt} = -(U - U_{\text{reset}}) + \Delta U
\end{equation}
where $\tau_m$ is a membrane time constant to control how fast the membrane voltage decays. $\Delta U$ and $U$ are the pre-synaptic input and membrane voltage value of the LIF neuron.

\section{Motivation} 
As a special form of sequential data, dynamic graphs exhibit non-negligible temporal dependencies at both the node and structure levels. Therefore, to guarantee the effectiveness of a condensed dynamic graph $\mathcal{S}$ when fed into a DGNN, it is essential to enforce temporal consistency across graph snapshots.  In other words, the joint distribution of $\mathcal{S}$ over the entire sequence of graph snapshots should align with that of $\mathcal{T}$. However, existing GC methods typically can only align the marginal distributions of individual snapshots, treating each time step independently and failing to capture global temporal coherence. This limitation is formalized and further analyzed in Proposition~\ref{pro}.


\begin{proposition}
Given any two dynamic graphs $\mathcal{T} = \{G^t\}_{t=1}^T$ and $\mathcal{S} = \{\tilde{G}^t\}_{t=1}^T$ follow the distributions $P_{G^{1:T}}$ and $Q_{\tilde{G}^{1:T}}$, respectively. 
Assume that each graph snapshot $G^t$ and $\tilde{G}^t$ follows the marginal distribution $P_{G^t}$ and $Q_{\tilde{G}^t}$, respectively. Then the following relation holds:

\begin{equation}
\underbrace{\mathcal{D}_{\mathrm{KL}} \bigl(P_{G^{1:T}} \,\|\, Q_{\tilde{G}^{1:T}}\bigr)}_{\text{Joint Divergence}} 
\equiv \underbrace{\sum_{t=1}^T  \mathcal{D}_{\mathrm{KL}}\bigl(P_{G^t}\,\|\,Q_{\tilde G^t} \bigr)}_{\text{Marginal Divergence}}
\quad \text{iff } 
\begin{cases}
P_{G^{1:T}} = \prod_{t=1}^T P_{G^t}, \\
Q_{\tilde{G}^{1:T}} = \prod_{t=1}^T Q_{\tilde{G}^t}.
\end{cases}
\end{equation}
where $\mathcal{D}_{\mathrm{KL}}$ denotes the Kullback–Leibler divergence \cite{CKL}, and $\equiv$ indicates that the two expressions are identically equal.

\label{pro}
\end{proposition}

Proposition~\ref{pro} underscores that aligning the marginal distributions of individual graph snapshots is sufficient to guarantee joint distribution alignment over the entire dynamic graph sequence if and only if the snapshots are \textbf{temporally independent}. Obviously, such a strong condition rarely holds in practice, as a real-world dynamic graph always exhibit significant temporal dependencies. 
Consequently, when existing GC methods meet dynamic graphs, they show limited capability in aligning the complex spatiotemporal distributions, as they overlook critical temporal dependencies. 

Therefore, it is essential to design tailored condensation frameworks for dynamic graphs to preserve the intricate and coupled spatiotemporal dependencies. To this end, we propose DyGC, an innovative framework that explicitly and faithfully aligns the overall spatiotemporal distribution of the dynamic graph. We detail its design and components in the following section.

\section{Methodology}

\begin{figure*}[h]
  \centering
   \includegraphics[width=1.0\linewidth]{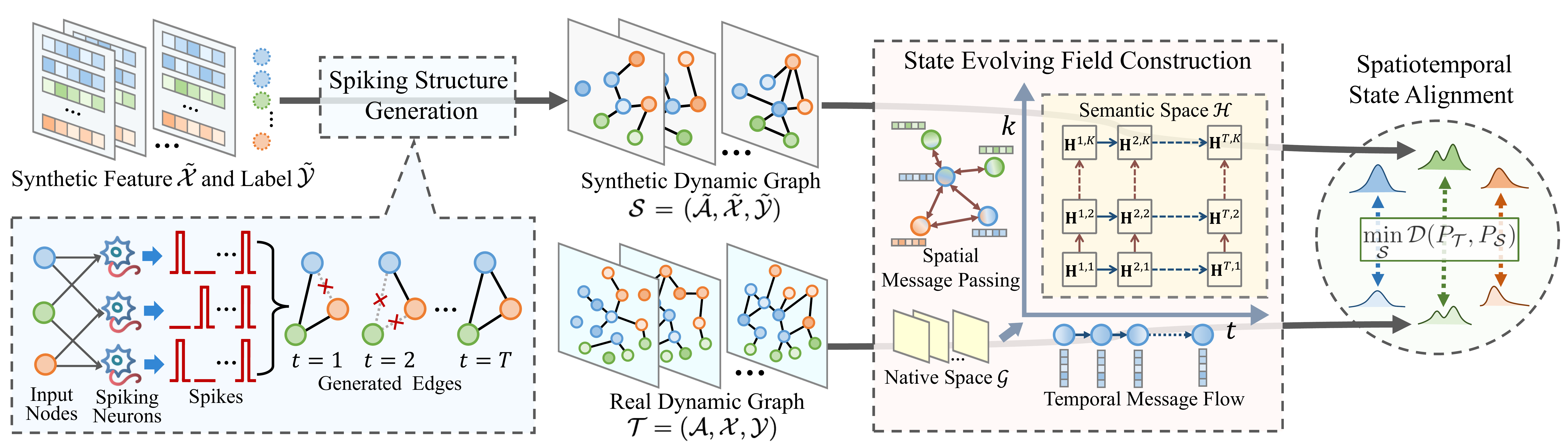}
   \caption{ Overall pipeline of the proposed Dynamic Graph Condensation (DyGC) framework. }
  \label{Framework}
\end{figure*}

The overall framework of the proposed DyGC is illustrated in Fig.~\ref{Framework}. Specifically, DyGC first introduces a spiking structure generation mechanism to synthesize an evolving and discrete graph structure $\tilde{\mathcal{A}}$ for $\mathcal{S}$ based on its synthetic node features $\tilde{\mathcal{X}}$. Subsequently, both real dynamic graph $\mathcal{T}$ and synthetic dynamic graph $\mathcal{S}$ are mapped into the semantic space $\mathcal{H}$ by constructing a state evolution field, which preserves their inherent spatiotemporal characteristics. A fine-grained spatiotemporal state alignment is then applied to optimize $\mathcal{S}$ such that it closely approximates the overall distribution of $\mathcal{T}$. Moreover, to improve the downstream performance of condensed graphs, we incorporate logit alignment to achieve task-oriented alignment optimization.

\subsection{Spiking Structure Generation}
During structure generation, we novelly leverage  the temporal activity of spiking neurons to model the interactions between nodes in dynamic graphs, and subsequently utilize their discrete spike outputs to induce the inter-node connectivity. In detail, for a node pair $(v_i, v_j) \in \mathcal{S}$ at time step $t$, their connectivity $\mathbf{\tilde{A}}^{(t)}_{ij}$ is jointly determined by both historical interaction patterns and current stimulus. This can be characterized as the iterative integration and firing dynamics in the LIF model:
\begin{equation}
 \textbf{Integrate:} \quad \hat{U}_{ij}^{(t)} =  U_{ij}^{(t-1)} + {\tau_m} \Big( \underbrace{U_{\text{reset}} - U_{ij}^{(t-1)}}_{\text{Memory Leakage}} +  \underbrace{\Psi(x_i^{(t)}, x_j^{(t)})}_{\text{Synaptic Input}} \Big),
\end{equation}
where the membrane voltage $U^{(t-1)}_{i,j}$ encodes the cumulative interaction memory between nodes up to time step $t-1$. The synaptic input of neurons is computed by the function $\psi(\cdot, \cdot)$, which measures current stimulus based on node features, such as with dot product, learnable linear layers, etc. The learnable decay factor $\tau_m$ modulates the influence of past interactions on current integration. Then, we have the inter-node affinity \( \hat{U}_{ij}^{(t)} \), which incorporates both historical and current affinity, and can be used to determine inter-node connectivity through the LIF neuron's firing behavior:
\begin{equation}
 \textbf{Fire:} \quad\mathbf{\tilde{A}}^{(t)}_{ij} =  \Theta (\hat{U}^{(t)}_{ij} - U^{(t)}_{\text{th}}),
\end{equation}
where $\Theta(\cdot)$ represents the Heaviside step function used to characterize the generation of edges, defined as $\Theta(x) = 1$ if $x \geq 0$ and 0 otherwise. The firing threshold $U^{(t)}_{\text{th}}$, which represents the minimum affinity required for a connection ($\mathbf{\tilde{A}}^{(t)}_{ij} = 1$), can be configured as a learnable parameter to accommodate dynamic graphs with varying densities. To avoid unbounded membrane voltage accumulation, a soft reset mechanism is introduced as $U_{ij}^{(t)} = \tilde{U}_{ij}^{(t)} - \tau_m \mathbf{\hat{A}}_{ij}^{(t)} (\hat{U}_{ij}^{(t)} - U_{\text{reset}}).$  This mechanism enables timely resetting of inter-node affinity based on spike outputs, thereby preventing topological rigidity in the evolving structure.

After the complete synthesized node feature sequence $\tilde{\mathcal{X}}$ is sequentially fed into spiking neurons, an evolving and discrete (without additional edge weights) structure $\tilde{\mathcal{A}}$ is defined based on binary spike outputs. Therefore, we obtain the structured $\mathcal{S}$ for subsequent distribution matching.

\subsection{Dynamic Graph Distribution Matching} \label{DGMP}
As shown in Eq.~\ref{opt}, the quality of the condensed graph $\mathcal{S}$ heavily depends on the objective of dynamic graph distribution matching. In DyGC, we achieve this matching by first constructing a semantically rich and tractable state evolving field for dynamic graphs. After that, a fine-grained spatiotemporal state alignment mechanism is employed to effectively guide the optimization of the synthetic graph.

\textbf{State Evolving Field for Dynamic Graphs.} To preserve the original spatiotemporal characteristics, we introduce a novel and  expressive state evolving field, constructed from the native space of dynamic graphs.  Specifically, this field characterizes the evolving states of dynamic graphs through joint message propagation across both spatial and temporal dimensions. Along the spatial dimension, a low-cost message passing scheme enables nodes to aggregate information from their local neighborhoods, facilitating efficient structural encoding. In contrast, temporal message flow establishes explicit interactions across successive graph snapshots to maintain node-level temporal consistency. 


Given a dynamic graph with node feature sequence $\mathcal{X}$ and evolving adjacency matrices $\mathcal{A}$, we define its state evolving field as
$\mathcal{H} = [ \mathbf{H}^{(t,k)} ]_{t=1,k=1}^{T,K} \in \mathbb{R}^{K \times T \times n \times d}$,
where $\mathbf{H}^{(t,k)}$ encodes the state of all nodes at spatial step $k$ and time step $t$, with 
$K$ denoting the total number of spatial propagation steps. Notably, each state explicitly depends on preceding temporal and spatial states and is recursively computed as follows:
\begin{equation}
\mathbf{H}^{(t,k)} =  \alpha \mathbf{H}^{(t-1,k)} + (1- \alpha) \mathbf{M}^{(t)} \mathbf{H}^{(t,k-1)}, \quad \text{for } 2 \leq t \leq T, \ 2 \leq k \leq K.
\end{equation}
Here, $\alpha \in [0,1]$ is a temporal propagation coefficient that governs the flow of message across the time dimension. At each timestep $t$, spatial message passing is performed through the state transition matrix $\mathbf{M}^{(t)} = \hat{\mathbf{D}}^{-\frac{1}{2}}\hat{\mathbf{A}}^{(t)}\hat{\mathbf{D}}^{-\frac{1}{2}} \in \mathbb{R}^{n \times n}$, where $\hat{\mathbf{A}}^{(t)} = \mathbf{A}^{(t)} + \mathbf{I}_n$ is the adjacency matrix with added self-loops and $\hat{\mathbf{D}}$ is the corresponding diagonal degree matrix.  Specifically, the boundary states in $\mathcal{H}$ are obtained through propagation along only a single dimension. The spatial boundary states are initialized as $\mathbf{H}^{(1,k)} = (\mathbf{M}^{(1)})^{k-1} \mathbf{X}^{(1)}$ for $k \in \{1, \dots, K\}$, while the temporal boundary states are computed as $\mathbf{H}^{(t,1)} = (1 - \alpha) \mathbf{X}^{(t)} + \alpha \mathbf{X}^{(t-1)}$ for $t \in \{2, \dots, T\}$.


Therefore, the real graph $\mathcal{T}$ and synthetic graph $\mathcal{S}$ are mapped to state evolving fields $\mathcal{H}^{\mathcal{T}}$ and $\mathcal{H}^{\mathcal{S}}$, which resides in an easily measurable semantic space.


\textbf{Spatiotemporal State Alignment.}  Through the state evolving field, we reformulate the dynamic graph distribution matching objective as alignment of spatiotemporal states in semantic space $\mathcal{H}$. However, existing GC methods typically focus on aligning low-order moments (e.g., mean and standard deviation), which is insufficient in dynamic settings. This limitation arises because higher-order statistics are essential for enhancing the distributional discriminability across different spatiotemporal states, thereby guiding the condensed graph to better reconstruct original evolution characteristics. Nevertheless, 
the explicit estimation of higher-order moments is non-trivial and cumbersome. Therefore, to address the limitation mentioned above,
Maximum Mean Discrepancy (MMD) \cite{MMD} provides a powerful solution by leveraging the kernel trick to implicitly compare all orders of moments. 


By leveraging MMD, we facilitate higher-order distribution alignment of spatiotemporal states in evolving fields $\mathcal{H}^{\mathcal{T}}$ and $\mathcal{H}^{\mathcal{S}}$. Moreover, to account for inter-class distributional variations,  we employ a class-wise strategy that separately estimates the discrepancy for each class of nodes. Let \( \mathcal{V}_c^{\mathcal{T}} = \{v_i \mid v_i \in \mathcal{V}_\mathcal{T}, y_i = c\} \) and \( \mathcal{V}_c^{\mathcal{S}} = \{v_i \mid v_i \in \mathcal{V}_\mathcal{S}, y_i = c\} \) denote the sets of nodes belonging to class \( c \) in the real and condensed graphs, respectively. Therefore, the objective of dynamic graph distribution matching in Eq.~\ref{opt} can be reformulated as:
\begin{equation}
\min_\mathcal{S} \underbrace{\sum_{c=1}^{C} \eta_{c} \left( 
 \mathcal{K}_c^{\mathcal{T}, \mathcal{T}}
+ \mathcal{K}_c^{\mathcal{S}, \mathcal{S}}
- 2\mathcal{K}_c^{\mathcal{T}, \mathcal{S}}
\right)}_{\text{Distribution Matching Loss } \mathcal{L}_{\text{dist}}}
\end{equation}
where \( \eta_c \) denotes the normalized proportion for class \( c \), controlling the alignment weight for each class. The kernel similarity term $\mathcal{K}_c^{{\mathcal{T}}, {\mathcal{S}}}$ is computed as:
\begin{equation}
\mathcal{K}_c^{\mathcal{T}, \mathcal{S}} = \frac{1}{|\mathcal{V}_c^{\mathcal{T}}| |\mathcal{V}_c^{\mathcal{S}}|}\sum_{v \in \mathcal{V}_c^{\mathcal{T}}} \sum_{u \in \mathcal{V}_c^{\mathcal{S}}} \psi(\mathcal{H}^{\mathcal{T}}_v, \mathcal{H}^{\mathcal{S}}_u),
\end{equation}
where $\mathcal{H}_v^{\mathcal{T}}$ and $\mathcal{H}_u^{\mathcal{S}} \in \mathbb{R}^{K \times T \times d}$ denote the tensor representations of node $v$ and $u$ extracted from their respective state evolving fields, and $\psi(\cdot)$ is the kernel function (e.g., RBF kernel or polynomial kernel). By minimizing these class-conditional discrepancies through $\mathcal{L}_{\text{dist}}$, the optimization guides $\mathcal{S}$ to faithfully preserve the spatiotemporal distributional properties of $\mathcal{T}$.

\subsection{Task-oriented Alignment Optimization}

To improve the performance of the condensed dynamic graph $\mathcal{S}$ on specific downstream tasks, it is essential to align high-level semantic information. Inspired by~\cite{SimGC}, we generalize Logits Alignment (LA) to dynamic graphs, ensuring that the condensation process yield a synthesized graph that is explicitly aligned with task-specific objectives. 
In detail, we first leverage a pre-trained DGNN on real graph to generate soft labels $\tilde{\mathcal{Y}}'$ using $\tilde{\mathcal{X}}$ and $\tilde{\mathcal{A}}$ as inputs. Then, we align these predicted logits with the synthesized labels by minimizing a classification-aware loss as:
$\mathcal{L}_{\text{logit}} =\sum_{i=1}^{m}-\tilde{\mathcal{Y}}_i\log \tilde{\mathcal{Y}}'_i$. Finally, combined with the distribution matching loss, the overall condensation objective of DyGC is computed as $\mathcal{L} = \mathcal{L}_{\text{dist}} + \gamma \mathcal{L}_{\text{logit}}$, where $\gamma$ is a trade-off hyperparameter.
Such a logits alignment optimization could implicitly capture the task-oriented knowledge from the real dynamic graph by learning the predictive behavior of the pre-trained DGNN. More detailed algorithmic descriptions of DyGC are provided in the Appendix \ref{algorithm}.

\section{Experiments} 

\subsection{Experimental Settings} 
\textbf{Datasets $\&$ Baseline Methods.} We conduct extensive evaluations on four real-world dynamic graph datasets, including DBLP, Reddit, Arxiv, and Tmall. These datasets vary in scale, domain, and temporal characteristics, providing  a comprehensive evaluation across a wide range of real-world scenarios. In addition, we adopt the following baselines for comprehensive comparisons: (1) Coreset-based methods, $i.e.$, Random, Herding~\cite{herding}, and K-Center~\cite{kcenter}; and (2) Static graph condensation methods, $i.e.$, GCond~\cite{GCond}, GCDM~\cite{GCDM}, SimGC~\cite{SimGC}, and CGC~\cite{CGC}. For static condensation methods, we apply them independently to each graph snapshot, and sequentially combine them to construct the entire dynamic graph.

\textbf{Implementations $\&$ Architectures for DGNNs.} The whole experimental pipeline is divided into two stages: (1) Condensation stage: synthesizing condensed dynamic graphs under varying condensation ratio; (2) Evaluation stage: training a certain DGNN model on the condensed dynamic graph obtained from the first stage, and evaluating it on the test set of the real graph. For evaluation stage, we focus on the temporal node classification task, in which the full graph structure at each time step is available. To further assess the generalization capability of DyGC, we evaluate  the condensed data using five representative DGNN backbone architectures: T-GCN~\cite{T-GCN}, GCRN~\cite{GCRN}, STGCN~\cite{STGCN}, DySAT~\cite{DySAT}, and ROLAND~\cite{ROLAND}. These backbones feature diverse combinations of static graph encoders and sequence modeling methods, representing a broad spectrum of design choices in dynamic graph learning. More detailed experimental settings are provided in Appendix~\ref{Experimental}.

\begin{table*}[t]
  \centering
  \caption{ Temporal node classification performance (\%) comparison under various condensation ratios. The \textcolor{new_red}{best} and the \textcolor{new_blue}{second best} results are highlighted as \textcolor{new_red}{\textbf{red}} and \textcolor{ new_blue}{\textbf{blue}}, respectively.}
  \vspace{0.5em}
      \resizebox{\textwidth}{!}
  {
    \begin{tabular}{ccc|cccccccc|c}
    \toprule
    \textbf{Dateset} & \textbf{Metric} & \textbf{Ratio ($r$)} & \textbf{Random} & \textbf{Herding} & \textbf{K-center} & \textbf{GCond} & \textbf{GCDM} & \textbf{SimGC} & \textbf{CGC} & \textbf{DyGC (Ours)} & \textbf{Whole} \\
    \midrule
    \multirow{6}[4]{*}{\textbf{DBLP}} & \multirow{3}[2]{*}{Micro-F1} & 0.5\% & 66.5{\scriptsize$\pm$3.4} & 68.9{\scriptsize$\pm$2.4} & 64.2{\scriptsize$\pm$1.7} & 64.2{\scriptsize$\pm$3.3} & 66.3{\scriptsize$\pm$2.5} & \textcolor{ new_blue}{\textbf{72.8{\scriptsize$\pm$1.1}}} & 65.5{\scriptsize$\pm$4.3} & \textcolor{new_red}{\textbf{77.1{\scriptsize$\pm$0.3}}} & \multirow{3}[2]{*}{79.2{\scriptsize$\pm$0.3}} \\
          &       & 2.5\% & 73.2{\scriptsize$\pm$1.7} & 66.7{\scriptsize$\pm$1.7} & 67.9{\scriptsize$\pm$1.1} & 69.7{\scriptsize$\pm$1.7} & 70.6{\scriptsize$\pm$2.8} & \textcolor{ new_blue}{\textbf{74.0{\scriptsize$\pm$0.8}}} & 72.3{\scriptsize$\pm$1.8} & \textcolor{new_red}{\textbf{78.0{\scriptsize$\pm$0.4}}} &  \\
          &       & 5.0\% & \textcolor{ new_blue}{\textbf{74.1{\scriptsize$\pm$1.1}}} & 65.2{\scriptsize$\pm$1.2} & 68.9{\scriptsize$\pm$1.4} & 67.0{\scriptsize$\pm$2.3} & 71.5{\scriptsize$\pm$2.3} & 73.2{\scriptsize$\pm$1.2} & 72.9{\scriptsize$\pm$1.0} & \textcolor{new_red}{\textbf{78.2{\scriptsize$\pm$0.3}}} &  \\
\cmidrule{2-12}          & \multirow{3}[2]{*}{Macro-F1} & 0.5\% & 64.4{\scriptsize$\pm$4.6} & 68.1{\scriptsize$\pm$5.2} & 62.1{\scriptsize$\pm$4.0} & 63.5{\scriptsize$\pm$4.6} & 64.0{\scriptsize$\pm$3.1} & \textcolor{ new_blue}{\textbf{71.9{\scriptsize$\pm$0.7}}} & 64.8{\scriptsize$\pm$5.1} & \textcolor{new_red}{\textbf{76.7{\scriptsize$\pm$0.2}}} & \multirow{3}[2]{*}{78.5{\scriptsize$\pm$0.3}} \\
          &       & 2.5\% & 72.4{\scriptsize$\pm$1.8} & 67.0{\scriptsize$\pm$1.6} & 64.5{\scriptsize$\pm$1.1} & 68.4{\scriptsize$\pm$2.2} & 69.0{\scriptsize$\pm$3.4} & \textcolor{ new_blue}{\textbf{73.2{\scriptsize$\pm$0.9}}} & 72.2{\scriptsize$\pm$1.2} & \textcolor{new_red}{\textbf{77.1{\scriptsize$\pm$0.5}}} &  \\
          &       & 5.0\% & \textcolor{ new_blue}{\textbf{73.5{\scriptsize$\pm$1.1}}} & 65.9{\scriptsize$\pm$1.0} & 64.8{\scriptsize$\pm$1.3} & 65.4{\scriptsize$\pm$3.7} & 69.8{\scriptsize$\pm$2.7} & 72.7{\scriptsize$\pm$1.3} & 73.0{\scriptsize$\pm$0.8} & \textcolor{new_red}{\textbf{77.4{\scriptsize$\pm$0.2}}} &  \\
    \midrule
    \multirow{6}[4]{*}{\textbf{Reddit}} & \multirow{3}[2]{*}{Micro-F1} & 0.5\% & 32.7{\scriptsize$\pm$1.3} & \textcolor{ new_blue}{\textbf{37.1{\scriptsize$\pm$0.7}}} & 36.9{\scriptsize$\pm$1.2} & 27.1{\scriptsize$\pm$1.7} & 29.6{\scriptsize$\pm$2.2} & 30.3{\scriptsize$\pm$1.2} & 28.4{\scriptsize$\pm$2.3} & \textcolor{new_red}{\textbf{39.9{\scriptsize$\pm$1.2}}} & \multirow{3}[2]{*}{49.0{\scriptsize$\pm$0.6}} \\
          &       & 2.5\% & 37.5{\scriptsize$\pm$1.3} & 36.8{\scriptsize$\pm$1.5} & \textcolor{ new_blue}{\textbf{41.3{\scriptsize$\pm$0.8}}} & 28.8{\scriptsize$\pm$1.4} & 32.3{\scriptsize$\pm$0.9} & 30.8{\scriptsize$\pm$0.9} & 35.8{\scriptsize$\pm$0.9} & \textcolor{new_red}{\textbf{44.5{\scriptsize$\pm$0.8}}} &  \\
          &       & 5.0\% & 39.2{\scriptsize$\pm$0.9} & 41.3{\scriptsize$\pm$0.7} & \textcolor{ new_blue}{\textbf{42.2{\scriptsize$\pm$0.9}}} & 28.0{\scriptsize$\pm$2.0} & 32.7{\scriptsize$\pm$1.0} & 29.8{\scriptsize$\pm$1.4} & 38.9{\scriptsize$\pm$1.1} & \textcolor{new_red}{\textbf{46.1{\scriptsize$\pm$0.6}}} &  \\
\cmidrule{2-12}          & \multirow{3}[2]{*}{Macro-F1} & 0.5\% & 30.2{\scriptsize$\pm$1.0} & \textcolor{ new_blue}{\textbf{37.1{\scriptsize$\pm$1.3}}} & 34.5{\scriptsize$\pm$0.8} & 17.7{\scriptsize$\pm$3.6} & 23.5{\scriptsize$\pm$4.3} & 27.4{\scriptsize$\pm$1.7} & 22.9{\scriptsize$\pm$3.3} & \textcolor{new_red}{\textbf{39.9{\scriptsize$\pm$1.0}}} & \multirow{3}[2]{*}{48.8{\scriptsize$\pm$0.7}} \\
          &       & 2.5\% & 36.0{\scriptsize$\pm$2.3} & 35.6{\scriptsize$\pm$2.4} & \textcolor{ new_blue}{\textbf{41.5{\scriptsize$\pm$0.7}}} & 21.0{\scriptsize$\pm$4.1} & 27.6{\scriptsize$\pm$1.6} & 26.9{\scriptsize$\pm$2.5} & 30.5{\scriptsize$\pm$1.9} & \textcolor{new_red}{\textbf{43.7{\scriptsize$\pm$0.9}}} &  \\
          &       & 5.0\% & 38.9{\scriptsize$\pm$1.2} & 41.2{\scriptsize$\pm$0.6} & \textcolor{ new_blue}{\textbf{42.0{\scriptsize$\pm$1.1}}} & 20.4{\scriptsize$\pm$5.1} & 28.4{\scriptsize$\pm$1.4} & 24.9{\scriptsize$\pm$2.2} & 37.0{\scriptsize$\pm$2.9} & \textcolor{new_red}{\textbf{45.7{\scriptsize$\pm$0.6}}} &  \\
    \midrule
    \multirow{6}[4]{*}{\textbf{Arxiv}} & \multirow{3}[2]{*}{Micro-F1} & 0.05\% & 38.0{\scriptsize$\pm$2.8} & 51.9{\scriptsize$\pm$1.2} & 43.8{\scriptsize$\pm$1.6} & OOM   & 50.3{\scriptsize$\pm$1.3} & 58.8{\scriptsize$\pm$0.6} & \textcolor{ new_blue}{\textbf{59.2{\scriptsize$\pm$0.5}}} & \textcolor{new_red}{\textbf{62.3{\scriptsize$\pm$0.7}}} & \multirow{3}[2]{*}{70.8{\scriptsize$\pm$0.2}} \\
          &       & 0.25\% & 52.7{\scriptsize$\pm$1.4} & 57.5{\scriptsize$\pm$0.7} & 49.4{\scriptsize$\pm$0.7} & OOM   & 57.9{\scriptsize$\pm$1.8} & \textcolor{ new_blue}{\textbf{62.7{\scriptsize$\pm$0.9}}} & 62.1{\scriptsize$\pm$0.7} & \textcolor{new_red}{\textbf{67.0{\scriptsize$\pm$0.4}}} &  \\
          &       & 0.50\% & 56.3{\scriptsize$\pm$1.3} & 57.9{\scriptsize$\pm$0.5} & 50.4{\scriptsize$\pm$1.4} & OOM   & 60.6{\scriptsize$\pm$0.5} & \textcolor{ new_blue}{\textbf{63.1{\scriptsize$\pm$0.7}}} & 62.5{\scriptsize$\pm$0.8} & \textcolor{new_red}{\textbf{68.1{\scriptsize$\pm$0.3}}} &  \\
\cmidrule{2-12}          & \multirow{3}[2]{*}{Macro-F1} & 0.05\% & 10.3{\scriptsize$\pm$1.6} & 18.2{\scriptsize$\pm$0.4} & 13.4{\scriptsize$\pm$0.6} & OOM   & 16.7{\scriptsize$\pm$1.5} & 29.3{\scriptsize$\pm$1.6} & \textcolor{ new_blue}{\textbf{31.9{\scriptsize$\pm$1.2}}} & \textcolor{new_red}{\textbf{34.2{\scriptsize$\pm$0.8}}} & \multirow{3}[2]{*}{49.9{\scriptsize$\pm$0.5}} \\
          &       & 0.25\% & 21.1{\scriptsize$\pm$1.0} & 30.1{\scriptsize$\pm$0.9} & 18.3{\scriptsize$\pm$0.4} & OOM   & 26.7{\scriptsize$\pm$1.5} & \textcolor{ new_blue}{\textbf{37.5{\scriptsize$\pm$0.7}}} & 35.7{\scriptsize$\pm$0.9} & \textcolor{new_red}{\textbf{41.1{\scriptsize$\pm$0.6}}} &  \\
          &       & 0.50\% & 25.0{\scriptsize$\pm$0.9} & 33.0{\scriptsize$\pm$0.4} & 22.2{\scriptsize$\pm$0.7} & OOM   & 30.4{\scriptsize$\pm$1.0} & \textcolor{ new_blue}{\textbf{39.5{\scriptsize$\pm$0.7}}} & 36.3{\scriptsize$\pm$1.8} & \textcolor{new_red}{\textbf{43.3{\scriptsize$\pm$0.4}}} &  \\
    \midrule
    \multirow{6}[4]{*}{\textbf{Tmall}} & \multirow{3}[2]{*}{Micro-F1} & 0.02\% & 43.9{\scriptsize$\pm$1.6} & 50.6{\scriptsize$\pm$1.0} & 48.7{\scriptsize$\pm$1.3} & OOM   & 49.6{\scriptsize$\pm$1.2} & \textcolor{ new_blue}{\textbf{55.6{\scriptsize$\pm$0.5}}} & 52.4{\scriptsize$\pm$1.9} & \textcolor{new_red}{\textbf{57.8{\scriptsize$\pm$0.8}}} & \multirow{3}[2]{*}{64.9{\scriptsize$\pm$0.3}} \\
          &       & 0.10\% & 51.1{\scriptsize$\pm$1.0} & 53.8{\scriptsize$\pm$0.6} & 54.9{\scriptsize$\pm$0.5} & OOM   & 48.5{\scriptsize$\pm$1.2} & \textcolor{ new_blue}{\textbf{57.3{\scriptsize$\pm$1.1}}} & 55.8{\scriptsize$\pm$0.7} & \textcolor{new_red}{\textbf{60.1{\scriptsize$\pm$0.5}}} &  \\
          &       & 0.20\% & 53.5{\scriptsize$\pm$0.4} & 55.4{\scriptsize$\pm$0.6} & 56.3{\scriptsize$\pm$0.4} & OOM   & 52.6{\scriptsize$\pm$0.9} & \textcolor{ new_blue}{\textbf{57.5{\scriptsize$\pm$0.9}}} & 54.7{\scriptsize$\pm$0.3} & \textcolor{new_red}{\textbf{60.5{\scriptsize$\pm$0.3}}} &  \\
\cmidrule{2-12}          & \multirow{3}[2]{*}{Macro-F1} & 0.02\% & 33.7{\scriptsize$\pm$2.2} & 44.0{\scriptsize$\pm$0.3} & 36.7{\scriptsize$\pm$2.2} & OOM   & 37.6{\scriptsize$\pm$2.4} & \textcolor{ new_blue}{\textbf{48.7{\scriptsize$\pm$1.4}}} & 46.2{\scriptsize$\pm$2.2} & \textcolor{new_red}{\textbf{50.9{\scriptsize$\pm$1.0}}} & \multirow{3}[2]{*}{59.8{\scriptsize$\pm$0.6}} \\
          &       & 0.10\% & 43.1{\scriptsize$\pm$1.3} & 48.4{\scriptsize$\pm$0.7} & 47.4{\scriptsize$\pm$0.6} & OOM   & 39.3{\scriptsize$\pm$1.8} & \textcolor{ new_blue}{\textbf{51.2{\scriptsize$\pm$1.3}}} & 49.8{\scriptsize$\pm$1.0} & \textcolor{new_red}{\textbf{54.0{\scriptsize$\pm$0.8}}} &  \\
          &       & 0.20\% & 46.6{\scriptsize$\pm$1.3} & 49.7{\scriptsize$\pm$0.6} & 49.8{\scriptsize$\pm$0.2} & OOM   & 45.4{\scriptsize$\pm$1.7} & \textcolor{ new_blue}{\textbf{51.9{\scriptsize$\pm$1.4}}} & 48.1{\scriptsize$\pm$1.5} & \textcolor{new_red}{\textbf{55.2{\scriptsize$\pm$0.3}}}
          &  \\
    \bottomrule
    \end{tabular}%
    }
  \label{com}%
\end{table*}%

\subsection{Experimental Results} 

\begin{wrapfigure}{r}{0.44\linewidth}
    \centering
    \vspace{-1mm}
    \includegraphics[width=\linewidth]{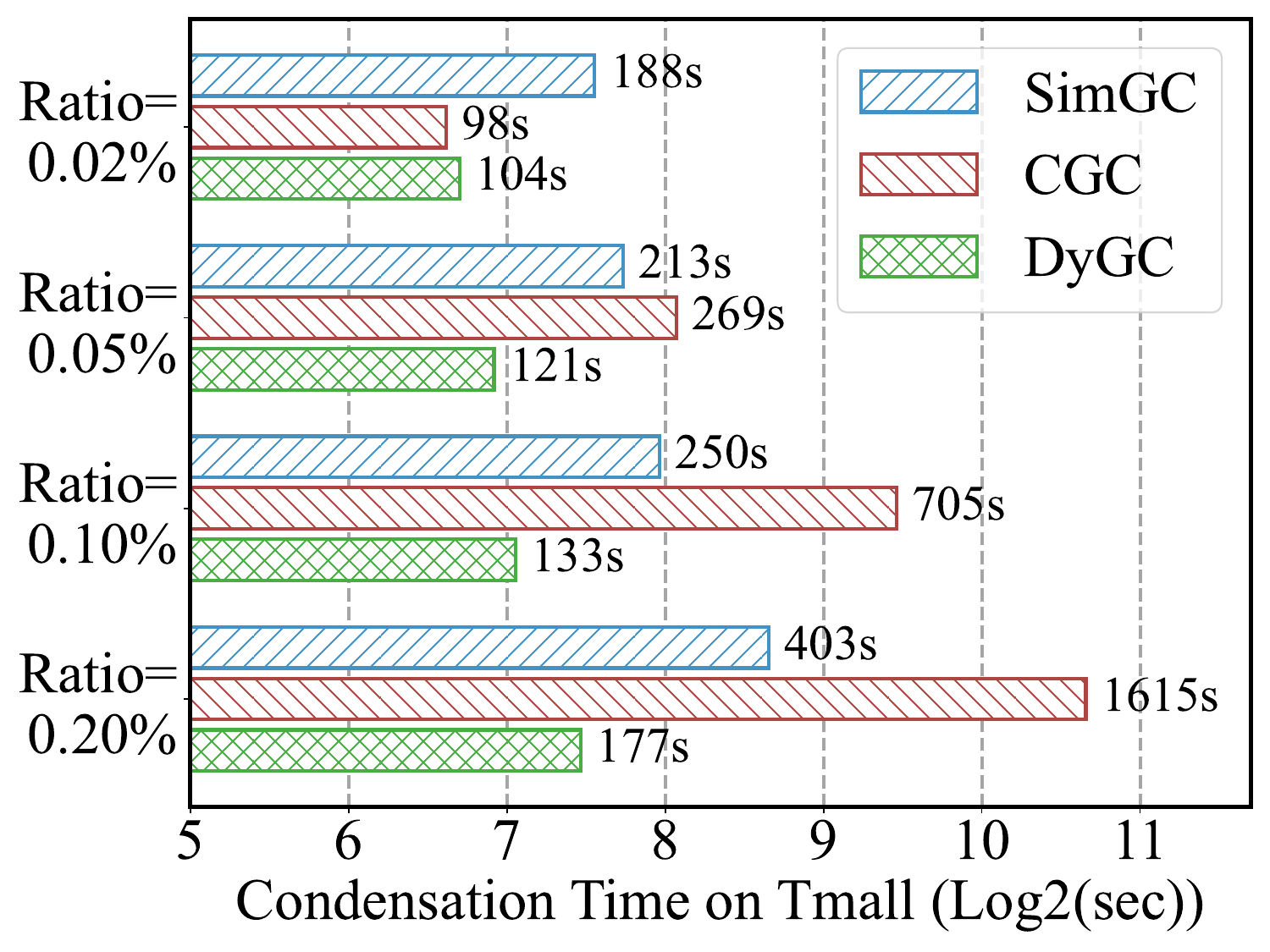}
    \caption{The time complexity comparison of different graph condensation methods.}  
    \vspace{-2mm}
    \label{condt}
\end{wrapfigure}

\textbf{Comparison with Baseline Methods.} We evaluate the temporal node classification performance of DyGC in comparison to other graph condensation (GC) methods, using T-GCN as a unified DGNN architecture.  The overall results are summarized in Table~\ref{com}. Across all settings (four datasets and three condensation ratios), DyGC consistently outperforms all baselines, achieving a substantial performance gain of 2.2\% to 5.0\%. This highlights the superior expressiveness of the condensed dynamic graphs generated by DyGC. This performance benefits from DyGC’s tailored strategies that thoroughly preserve the spatiotemporal characteristics of real dynamic graphs, ensuring better suitability for DGNN training. In contrast, static GC methods ignore temporal dependencies across graph snapshots, resulting in degraded performance. As for coreset-based methods,  which select only a small subset of nodes, struggle to adequately capture the full distribution of the dynamic graph. Notably, on DBLP and Reddit, static GC methods even underperform the coreset baselines in some cases. This is largely due to their disregard for temporal continuity, which highlights the importance of preserving temporally consistent evolutionary properties within dynamic graphs.

Meanwhile, DyGC exhibits significantly higher condensation efficiency. As shown in Fig.~\ref{condt}, DyGC achieves lower condensation time than static GC methods in most cases. This improvement stems from its integrated and parallelized processing of the entire dynamic graph sequence, along with a model-free distribution matching (i.e., constructing and aligning the state evolution field) that imposes minimal computational overhead. DyGC’s condensation time also increases more gradually as the condensation ratio rises, demonstrating better scalability in optimizing condensed graphs across varying scale requirements.

\begin{table*}[t]
  \centering
  \caption{Performance (\%) and fidelity across different DGNN architectures.}
        \resizebox{\textwidth}{!}
  {
        \begin{tabular}{cc|cccccccccc}
    \toprule
    \multirow{2}[4]{*}{\textbf{Dataset}} & \multicolumn{1}{c}{\multirow{2}[4]{*}{\textbf{Ratio}}} & \multicolumn{2}{c}{\textbf{T-GCN }} & \multicolumn{2}{c}{\textbf{GCRN}} & \multicolumn{2}{c}{\textbf{STGCN }} & \multicolumn{2}{c}{\textbf{DySAT}} & \multicolumn{2}{c}{\textbf{ROLAND}} \\
\cmidrule{3-12}          & \multicolumn{1}{c}{} & Micro-F1 & Macro-F1 & Micro-F1 & Macro-F1 & Micro-F1 & Macro-F1 & Micro-F1 & Macro-F1 & Micro-F1 & Macro-F1 \\
    \midrule
    \multirow{4}[4]{*}{\textbf{DBLP}} & Whole & 79.2{\scriptsize$\pm$0.3} & 78.5{\scriptsize$\pm$0.3} & 77.6{\scriptsize$\pm$0.4} & 77.2{\scriptsize$\pm$0.4} & 76.6{\scriptsize$\pm$0.7} & 75.8{\scriptsize$\pm$0.6} & 78.8{\scriptsize$\pm$0.3} & 78.3{\scriptsize$\pm$0.3} & 79.4{\scriptsize$\pm$0.4} & 79.1{\scriptsize$\pm$0.3} \\
          & 0.5\% & 77.1{\scriptsize$\pm$0.3} & 76.7{\scriptsize$\pm$0.2} & 72.5{\scriptsize$\pm$1.6} & 71.7{\scriptsize$\pm$1.4} & 72.5{\scriptsize$\pm$1.6} & 71.7{\scriptsize$\pm$1.4} & 76.5{\scriptsize$\pm$0.5} & 76.0{\scriptsize$\pm$0.1} & 76.6{\scriptsize$\pm$0.9} & 76.2{\scriptsize$\pm$0.7} \\
          & 5.0\% & \textbf{78.2{\scriptsize$\pm$0.3}} & \textbf{77.4{\scriptsize$\pm$0.5}} & \textbf{75.2{\scriptsize$\pm$0.6}} & \textbf{74.5{\scriptsize$\pm$0.4}} & \textbf{75.2{\scriptsize$\pm$0.6}} & \textbf{74.5{\scriptsize$\pm$0.4}} & \textbf{77.2{\scriptsize$\pm$0.4}} & \textbf{76.8{\scriptsize$\pm$0.3}} & \textbf{78.4{\scriptsize$\pm$0.6}} & \textbf{77.7{\scriptsize$\pm$0.5}} \\
\cmidrule{2-12}          & \cellcolor[rgb]{.922, 1, .925}\textbf{Fidelity } & \cellcolor[rgb]{.922, 1, .925}\textbf{98.7\%} & \cellcolor[rgb]{.922, 1, .925}\textbf{98.6\%} & \cellcolor[rgb]{.922, 1, .925}\textbf{96.9\%} & \cellcolor[rgb]{.922, 1, .925}\textbf{96.5\%} & \cellcolor[rgb]{.922, 1, .925}\textbf{98.2\%} & \cellcolor[rgb]{.922, 1, .925}\textbf{98.3\%} & \cellcolor[rgb]{.922, 1, .925}\textbf{98.0\%} & \cellcolor[rgb]{.922, 1, .925}\textbf{98.1\%} & \cellcolor[rgb]{.922, 1, .925}\textbf{98.7\%} & \cellcolor[rgb]{.922, 1, .925}\textbf{98.2\%} \\
    \midrule
    \multirow{4}[4]{*}{\textbf{Reddit}} & Whole & 49.0{\scriptsize$\pm$0.6} & 48.8{\scriptsize$\pm$0.7} & 45.0{\scriptsize$\pm$1.2} & 42.8{\scriptsize$\pm$2.2} & 46.2{\scriptsize$\pm$0.7} & 45.7{\scriptsize$\pm$1.0} & 39.5{\scriptsize$\pm$1.3} & 38.1{\scriptsize$\pm$0.9} & 49.6{\scriptsize$\pm$0.6} & 49.6{\scriptsize$\pm$0.6} \\
          & 0.5\% & 39.9{\scriptsize$\pm$1.2} & 39.9{\scriptsize$\pm$1.0} & 36.1{\scriptsize$\pm$2.0} & 35.4{\scriptsize$\pm$3.2} & 36.7{\scriptsize$\pm$1.1} & 36.3{\scriptsize$\pm$0.4} & 32.9{\scriptsize$\pm$1.6} & 32.2{\scriptsize$\pm$1.2} & 38.5{\scriptsize$\pm$1.4} & 37.6{\scriptsize$\pm$0.6} \\
          & 5.0\% & \textbf{46.1{\scriptsize$\pm$0.6}} & \textbf{45.7{\scriptsize$\pm$0.6}} & \textbf{40.5{\scriptsize$\pm$2.1}} & \textbf{38.4{\scriptsize$\pm$3.0}} & \textbf{42.1{\scriptsize$\pm$0.5}} & \textbf{41.8{\scriptsize$\pm$0.4}} & \textbf{37.0{\scriptsize$\pm$1.0}} & \textbf{36.1{\scriptsize$\pm$1.5}} & \textbf{46.2{\scriptsize$\pm$0.8}} & \textbf{45.7{\scriptsize$\pm$0.9}} \\
\cmidrule{2-12}          & \cellcolor[rgb]{.922, 1, .925}\textbf{Fidelity} & \cellcolor[rgb]{.922, 1, .925}\textbf{94.1\%} & \cellcolor[rgb]{.922, 1, .925}\textbf{93.6\%} & \cellcolor[rgb]{.922, 1, .925}\textbf{90.0\%} & \cellcolor[rgb]{.922, 1, .925}\textbf{89.7\%} & \cellcolor[rgb]{.922, 1, .925}\textbf{91.1\%} & \cellcolor[rgb]{.922, 1, .925}\textbf{91.5\%} & \cellcolor[rgb]{.922, 1, .925}\textbf{93.7\%} & \cellcolor[rgb]{.922, 1, .925}\textbf{94.8\%} & \cellcolor[rgb]{.922, 1, .925}\textbf{93.1\%} & \cellcolor[rgb]{.922, 1, .925}\textbf{92.1\%} \\
    \midrule
    \multirow{4}[4]{*}{\textbf{Arxiv}} & Whole & 70.8{\scriptsize$\pm$0.2} & 49.9{\scriptsize$\pm$0.5} & 65.9{\scriptsize$\pm$0.1} & 43.1{\scriptsize$\pm$0.1} & 66.6{\scriptsize$\pm$0.1} & 43.8{\scriptsize$\pm$0.2} & 71.2{\scriptsize$\pm$0.2} & 50.9{\scriptsize$\pm$0.1} & 69.8{\scriptsize$\pm$0.6} & 49.5{\scriptsize$\pm$0.8} \\
          & 0.05\% & 62.3{\scriptsize$\pm$0.7} & 34.2{\scriptsize$\pm$0.8} & 56.4{\scriptsize$\pm$0.8} & 32.9{\scriptsize$\pm$1.1} & 57.6{\scriptsize$\pm$0.4} & 34.9{\scriptsize$\pm$0.2} & 62.6{\scriptsize$\pm$1.1} & 38.7{\scriptsize$\pm$0.4} & 59.9{\scriptsize$\pm$0.4} & 34.8{\scriptsize$\pm$1.2} \\
          & 0.50\% & \textbf{68.1{\scriptsize$\pm$0.3}} & \textbf{43.3{\scriptsize$\pm$0.4}} & \textbf{63.5{\scriptsize$\pm$0.4}} & \textbf{37.9{\scriptsize$\pm$0.7}} & \textbf{63.9{\scriptsize$\pm$0.4}} & \textbf{38.8{\scriptsize$\pm$0.6}} & \textbf{67.1{\scriptsize$\pm$0.1}} & \textbf{44.0{\scriptsize$\pm$0.6}} & \textbf{65.1{\scriptsize$\pm$0.4}} & \textbf{42.2{\scriptsize$\pm$1.6}} \\
\cmidrule{2-12}          & \cellcolor[rgb]{.922, 1, .925}\textbf{Fidelity} & \cellcolor[rgb]{.922, 1, .925}\textbf{96.2\%} & \cellcolor[rgb]{.922, 1, .925}\textbf{86.8\%} & \cellcolor[rgb]{.922, 1, .925}\textbf{96.4\%} & \cellcolor[rgb]{.922, 1, .925}\textbf{87.9\%} & \cellcolor[rgb]{.922, 1, .925}\textbf{95.9\%} & \cellcolor[rgb]{.922, 1, .925}\textbf{88.6\%} & \cellcolor[rgb]{.922, 1, .925}\textbf{94.2\%} & \cellcolor[rgb]{.922, 1, .925}\textbf{86.4\%} & \cellcolor[rgb]{.922, 1, .925}\textbf{93.3\%} & \cellcolor[rgb]{.922, 1, .925}\textbf{85.3\%} \\
    \midrule
    \multirow{4}[4]{*}{\textbf{Tmall}} & Whole & 64.9{\scriptsize$\pm$0.3} & 59.8{\scriptsize$\pm$0.6} & 64.1{\scriptsize$\pm$0.1} & 57.9{\scriptsize$\pm$0.4} & 63.2{\scriptsize$\pm$0.1} & 58.1{\scriptsize$\pm$0.3} & 64.0{\scriptsize$\pm$0.2} & 59.1{\scriptsize$\pm$0.3} & 64.3{\scriptsize$\pm$0.2} & 59.1{\scriptsize$\pm$0.4} \\
          & 0.02\% & 57.8{\scriptsize$\pm$0.8} & 50.9{\scriptsize$\pm$1.0} & 54.3{\scriptsize$\pm$0.4} & 50.8{\scriptsize$\pm$0.8} & \textbf{59.8{\scriptsize$\pm$0.2}} & 53.6{\scriptsize$\pm$0.6} & 57.9{\scriptsize$\pm$0.4} & 49.2{\scriptsize$\pm$1.4} & 58.7{\scriptsize$\pm$0.2} & 51.4{\scriptsize$\pm$0.6} \\
          & 0.20\% & \textbf{60.5{\scriptsize$\pm$0.3}} & \textbf{55.2{\scriptsize$\pm$0.3}} & \textbf{57.1{\scriptsize$\pm$0.7}} & \textbf{54.2{\scriptsize$\pm$0.5}} & 59.7{\scriptsize$\pm$0.2} & \textbf{55.6{\scriptsize$\pm$0.1}} & \textbf{59.8{\scriptsize$\pm$0.4}} & \textbf{54.5{\scriptsize$\pm$1.1}} & \textbf{60.3{\scriptsize$\pm$0.2}} & \textbf{54.5{\scriptsize$\pm$0.4}} \\
\cmidrule{2-12}          & \cellcolor[rgb]{.922, 1, .925}\textbf{Fidelity} & \cellcolor[rgb]{.922, 1, .925}\textbf{93.2\%} & \cellcolor[rgb]{.922, 1, .925}\textbf{92.3\%} & \cellcolor[rgb]{.922, 1, .925}\textbf{89.1\%} & \cellcolor[rgb]{.922, 1, .925}\textbf{93.6\%} & \cellcolor[rgb]{.922, 1, .925}\textbf{94.5\%} & \cellcolor[rgb]{.922, 1, .925}\textbf{95.7\%} & \cellcolor[rgb]{.922, 1, .925}\textbf{93.4\%} & \cellcolor[rgb]{.922, 1, .925}\textbf{92.2\%} & \cellcolor[rgb]{.922, 1, .925}\textbf{93.8\%} & \cellcolor[rgb]{.922, 1, .925}\textbf{92.2\%} \\
    \bottomrule
    \end{tabular}%
    }
  \label{tab:addlabel}%
\end{table*}%
\begin{figure*}[t]
    \centering
    \begin{minipage}[b]{0.40\textwidth}
        \centering
        \includegraphics[width=\linewidth]{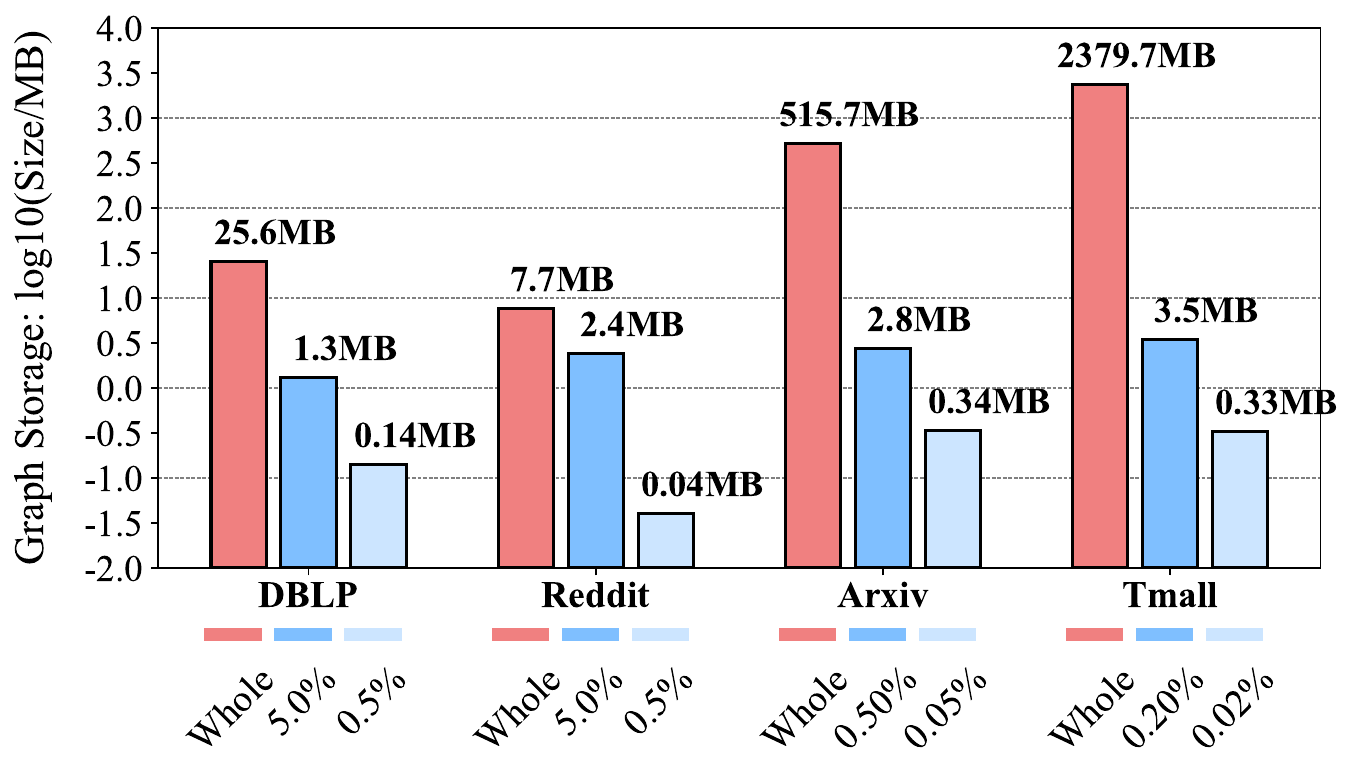}
        \label{fig:framework1}
    \end{minipage}
    \begin{minipage}[b]{0.58\textwidth}
        \centering
        \includegraphics[width=\linewidth]{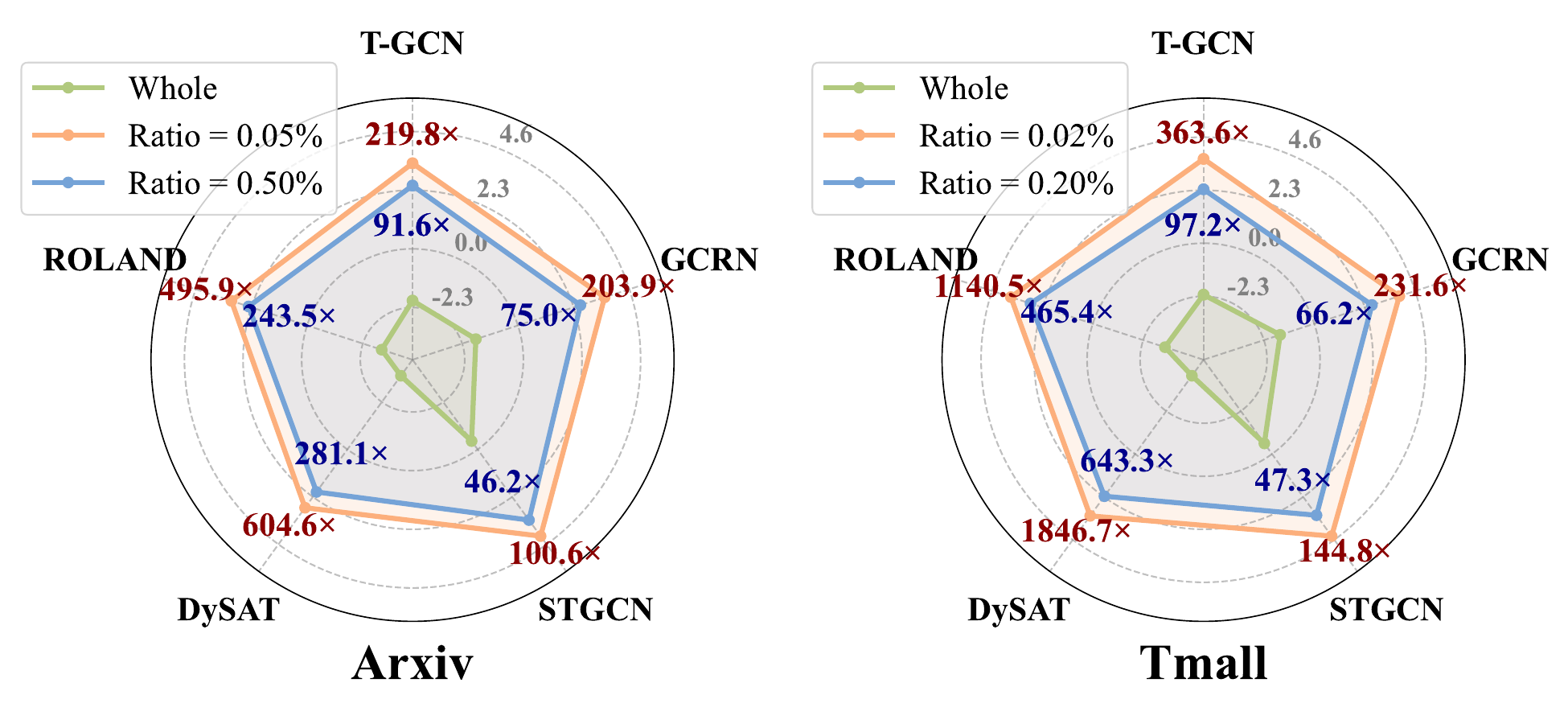}
        \label{fig:framework2}
    \end{minipage}
    \vspace{-0.5cm}
    \caption{Storage reduction and DGNN training speedup ($\text{Scale} = \ln\left(\text{epoch/sec}\right)$) via DyGC.}
    \label{effi}
\end{figure*}

\textbf{Generalization Ability of DyGC across Different DGNNs.} To assess the generalization ability of DyGC, we evaluate the performance of the condensed dynamic graph across five different DGNN architectures. As shown in our results, the condensed data consistently maintains high performance fidelity (86.4\%–98.7\%), regardless of the underlying DGNN model, demonstrating the strong generalization capability of DyGC. This robustness is attributed to DyGC’s use of spatiotemporal message propagation as a model-free condensation strategy, which enables it to capture the underlying distribution of dynamic graphs without being tied to the parameters of any specific DGNN. In addition, the incorporation of biologically inspired spiking structure generation introduces inherent noise robustness and temporal continuity, further enhancing the adaptability of the condensed graph.

\begin{wraptable}{r}{0.52\linewidth}
  \centering
  \caption{Ablation study of modules: spiking structure generation (SSG), dynamic graph distribution matching (DGM), and logits alignment (LA).}
  \resizebox{\linewidth}{!}{
    \begin{tabular}{ccc|cccc}
    \toprule
\multirow{2}{*}{\textbf{SSG}} & \multirow{2}{*}{\textbf{DGM}} & \multirow{2}{*}{\textbf{LA}} & \textbf{DBLP} & \textbf{Reddit} & \textbf{Arxiv} & \textbf{Tmall} \\
         &    &      & (2.5\%) & (2.5\%) & (0.25\%) & (0.10\%) \\
    \midrule
          &       &       & 57.8{\scriptsize$\pm$0.6} & 36.5{\scriptsize$\pm$1.2} & 33.4{\scriptsize$\pm$1.9} & 40.5{\scriptsize$\pm$1.0} \\
    \checkmark     &       &       & 61.4{\scriptsize$\pm$1.2} & 37.3{\scriptsize$\pm$0.8} & 39.7{\scriptsize$\pm$1.7} & 49.2{\scriptsize$\pm$1.5} \\
          & \checkmark     &       & 75.2{\scriptsize$\pm$0.7} & 42.6{\scriptsize$\pm$0.4} & 60.6{\scriptsize$\pm$0.9} & 56.0{\scriptsize$\pm$0.5} \\
          &       & \checkmark     & 63.3{\scriptsize$\pm$2.4} & 38.9{\scriptsize$\pm$0.7} & 64.5{\scriptsize$\pm$0.6} & 54.4{\scriptsize$\pm$0.8} \\
    \checkmark     & \checkmark     &       & 76.8{\scriptsize$\pm$0.5} & \underline{44.2{\scriptsize$\pm$0.8}} & 60.7{\scriptsize$\pm$0.5} & 55.6{\scriptsize$\pm$0.5} \\
    \checkmark     &       & \checkmark     & 69.5{\scriptsize$\pm$1.6} & 40.0{\scriptsize$\pm$1.4} & 63.5{\scriptsize$\pm$0.5} & 58.1{\scriptsize$\pm$0.2} \\
          & \checkmark     & \checkmark     & \underline{77.8{\scriptsize$\pm$0.2}} & 44.1{\scriptsize$\pm$0.9} & \underline{66.5{\scriptsize$\pm$0.7}} & \underline{59.7{\scriptsize$\pm$0.3}} \\
    \midrule
    \rowcolor[rgb]{.922, 1, .925} \checkmark     & \checkmark     & \checkmark     & \textbf{78.0{\scriptsize$\pm$0.4}} & \textbf{44.5{\scriptsize$\pm$0.8}} & \textbf{67.0{\scriptsize$\pm$0.4}} & \textbf{60.1{\scriptsize$\pm$0.3}} \\
    \bottomrule
    \end{tabular}%
    }
    \vspace{-1em}
  \label{Ablation}%
\end{wraptable}

\textbf{Efficiency Gains from Dynamic Graph Condensation.} To validate the efficiency gains introduced by DyGC, we compare the storage footprint and DGNN training speed of the condensed graphs against those of the original full graphs. As shown in Figure~\ref{effi}, the condensed graphs require substantially less storage. Notably, for the Tmall dataset, the storage size is reduced from 2.38GB to just 0.33MB under a condensation ratio of 0.02\%. In addition, the condensed graphs enable significant training acceleration across various DGNN architectures. For example, on the Tmall dataset, training DySAT—a representative attention-based DGNN—on the condensed graph (ratio = 0.02\%) yields a remarkable 1846.7× speedup. These results collectively demonstrate that the dynamic graphs synthesized by DyGC not only preserve the predictive performance of DGNNs but also deliver substantial improvements in computational efficiency, highlighting DyGC as a practical and scalable solution for dynamic graph learning.

\subsection{Ablation Study}
\textbf{Impact of Key Modules on Overall Performance (Micro-F1 \%). } As presented in Table~\ref{Ablation}, each component of DyGC plays a distinct and complementary role in enhancing condensation quality.  Notably, DGM consistently brings significant performance gains under all settings (yielding a 2.0\%-8.5\% improvement), demonstrating its critical role in aligning the spatiotemporal distribution. DGM leverages a tailored state evolving field to effectively encode both topology and temporal dependencies, enabling the condensed graph to closely approximate the original distribution.  LA demonstrates significant contributions on large-scale datasets such as Arxiv and Tmall (yielding a 4.5\%-6.3\% improvement). This highlights LA’s role in better adapting condensed dynamic graph to downstream tasks through effective alignment of high-level semantics. Although the performance gain from SSG is relatively modest, it explicitly establishes temporal dependencies on the topology through spiking neural encoding and generates discrete structures (without additional edge weights) via binary spiking behaviors. Such a biologically-inspired mechanism better reflects the evolving and event-driven nature of topology formation in real-world dynamic graphs.

\begin{figure*}[t]
    \centering
    \begin{minipage}[b]{0.60\textwidth}
        \centering
        \begin{subfigure}[b]{0.49\textwidth}
            \centering
            \includegraphics[width=\linewidth]{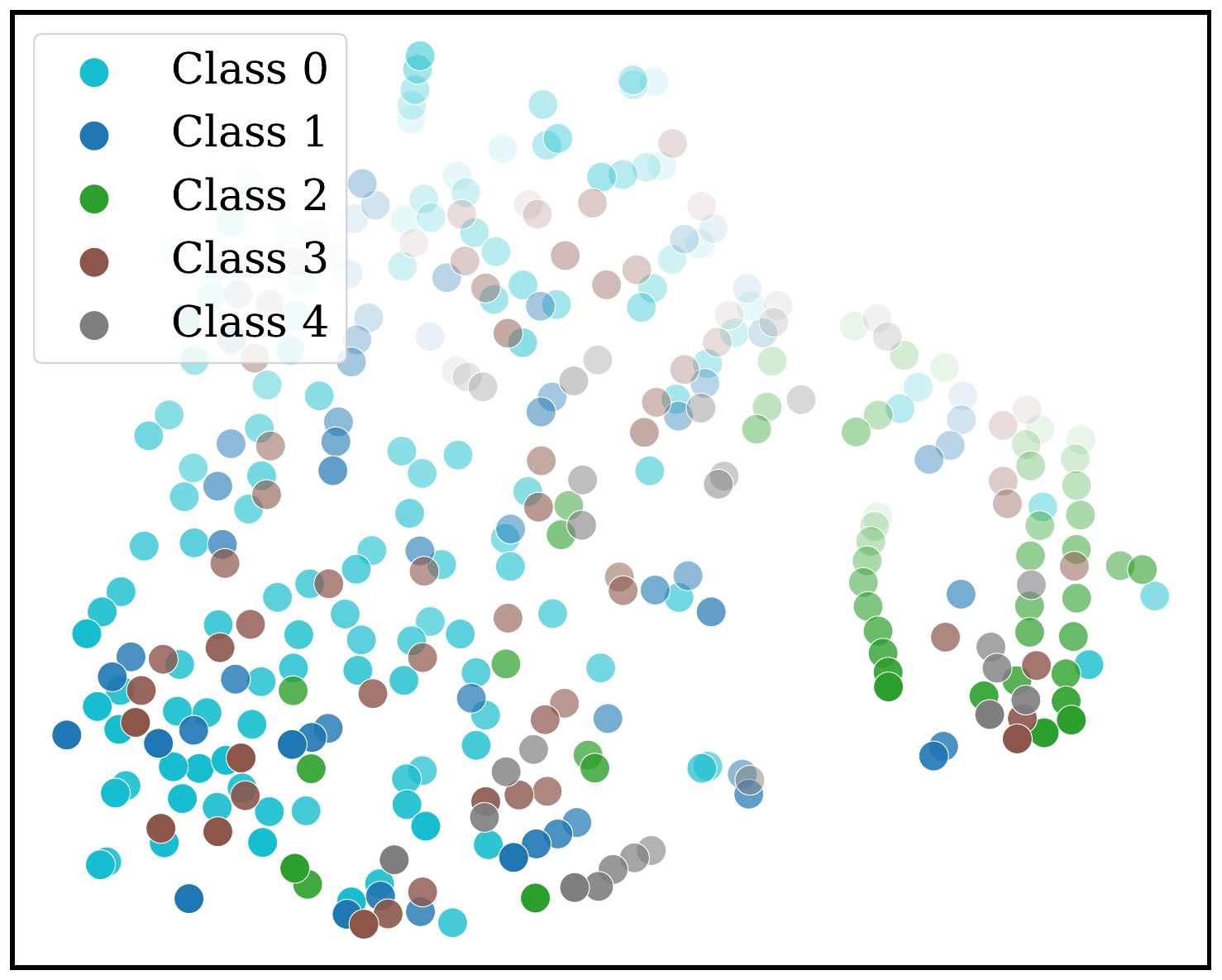}
            \caption{SimGC on DBLP.}
            \label{ss1}
        \end{subfigure}
        \hfill
        \begin{subfigure}[b]{0.49\textwidth}
            \centering
            \includegraphics[width=\linewidth]{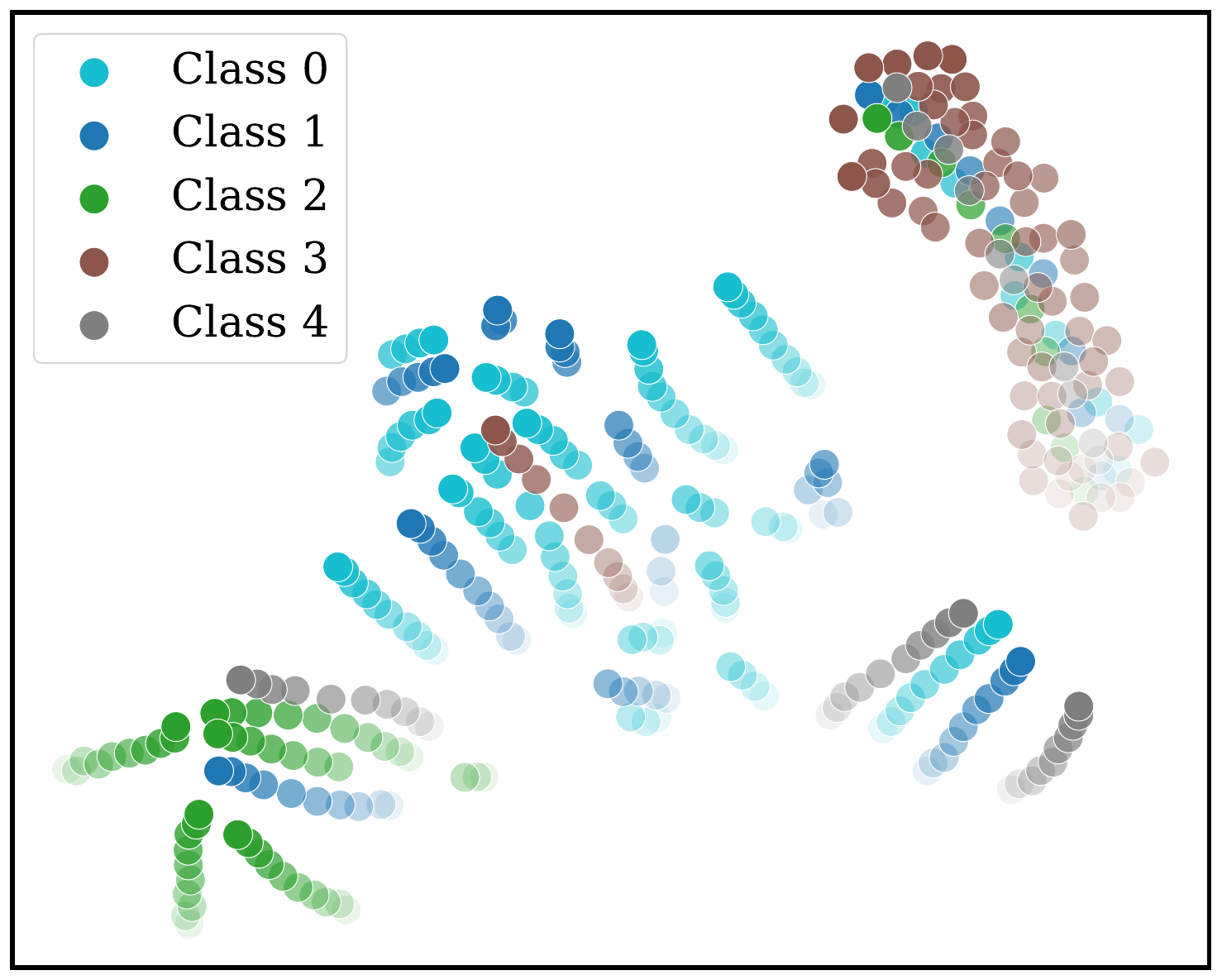}
            \caption{DyGC (Ours) on DBLP.}
            \label{ss2}
        \end{subfigure}
        \caption{t-SNE visualization of condensed node features. Point opacity from light to dark indicates temporal order.}
        \label{ntc}
    \end{minipage}
    \hfill
    \begin{minipage}[b]{0.38\textwidth}
        \centering
        \includegraphics[width=\linewidth]{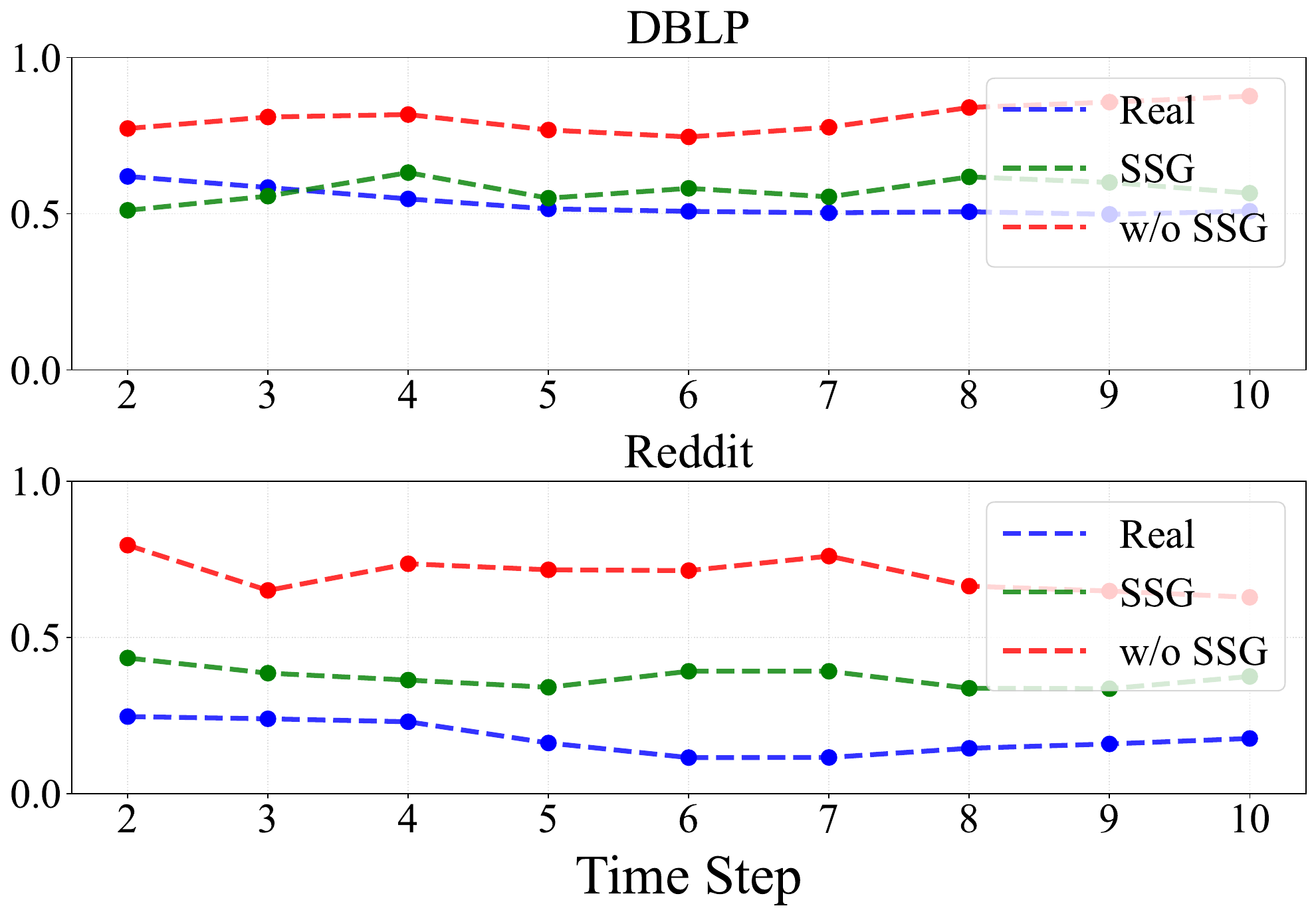}
        \caption{Temporal continuity of structures (Jaccard similarity) between consecutive time steps.}
        \label{stc}
    \end{minipage}
\end{figure*}

\textbf{Node-level Temporal Consistency}  To examine temporal coherence in the condensed graphs, we visualize the condensed node features at different time steps obtained using DyGC and SimGC. As illustrated in Figure~\ref{ntc}, DyGC generates condensed features with strong node-level temporal consistency. Specifically, each node at different time steps remain close in the feature space and show a clear temporal evolution trend. This property stems from its specialized distribution matching approach, which preserves the joint spatiotemporal distribution to maintain sequential integrity across graph snapshots. In contrast, SimGC aligns only the marginal distributions at each time step, which fails to ensure node-level temporal continuity and leads to disorder in the feature space.

\textbf{Approximate Temporal Continuity of Structures.} Structural continuity is critical for faithfully capturing the topological evolution of dynamic graphs. As shown in Figure~\ref{stc}, we quantify temporal structural continuity using the Jaccard similarity between consecutive graph snapshots. The results reveal that our spiking structure generation (SSG) module preserves a level of structural continuity similar to that of real graphs. By contrast, removing SSG leads to excessively rigid structures, as indicated by persistently high similarity between consecutive snapshots. This suggests a potential benefit of SSG in supporting more natural structural evolution over time.

\section{Conclusion}
In this work, we initiate the study of dynamic graph condensation and propose DyGC, a novel framework that condenses large-scale dynamic graphs into a compact yet informative representation for efficient DGNN training. DyGC incorporates a spiking structure generation mechanism to capture temporally evolving graph topologies and a tailored dynamic graph distribution matching strategy to preserve fine-grained spatiotemporal characteristics. Extensive experiments show that DyGC achieves substantial graph compression with minimal performance degradation, offering a scalable and effective data-centric solution for dynamic graph learning. In future work, we plan to extend DyGC to continuous-time dynamic scenarios and explore its applications in broader dynamic tasks, such as temporal link prediction, dynamic anomaly detection, and event-based reasoning, further enhancing its versatility and practical impact.

\newpage

\bibliographystyle{abbrv}
\bibliography{main}

\newpage

\appendix
\section*{Appendix}

This appendix supplements our work, \textbf{Dynamic Graph Condensation}, by providing additional details on the proposed DyGC. Specifically, it covers the broader impact, related work, proof of the proposition, algorithm descriptions, experimental setup with additional experiments and discussions.

\section{Broader Impact}
\label{appendix:broader_impact}
Our work on dynamic graph condensation may have broader societal impacts, particularly when applied to domains such as social networks, graph anomaly detection, and financial analytics where the underlying data is sensitive and evolving. Although our framework is designed as a fundamental graph learning tool without targeting specific applications, the ability to condense large-scale dynamic graphs raises concerns regarding potential misuse. For instance, improved condensation could inadvertently enable more efficient surveillance, by distilling temporal behavior patterns of individuals or groups into smaller, easily analyzable forms. Furthermore, if the condensed representations reflect inherent biases in the original data, they may reinforce unfair outcomes in downstream models, especially in high-stakes applications. We urge practitioners to be mindful of these implications, and to adopt safeguards such as fairness-aware learning, privacy-preserving data processing, and transparent evaluation protocols. As with all powerful data-centric methodologies, the deployment of dynamic graph condensation should prioritize ethical responsibility and social accountability.

\section{Related Works} \label{apprw}
\textbf{Graph Condensation.} As the pioneering GC methods, GCond~\cite{GCond} and DosCond~\cite{DosCond} introduced a gradient matching paradigm where the condensed graph is optimized by aligning GNN gradients computed on original and condensed graphs.  Building on this, SFGC~\cite{SFGC} innovatively distills graphs into a set of nodes without an explicit graph structure via training trajectory matching. However, these optimization-oriented methods necessitate GNN as a relay model, thereby introducing complex bi-level optimization problems~\cite{GCsurvey}. To avoid this,  distribution matching based methods such as GCDM~\cite{GCDM}, DisCo~\cite{Disco} and SimGC~\cite{SimGC} reformulate the matching objective as approximating the original graph in feature or representation space. Consequently, these methods obviates the expensive computation of gradients and trajectories in each training iteration of the relay model. Furthermore, CGC~\cite{CGC} proposes a training-free framework that transforms the GC optimization into a class partition problem which significantly enhances condensation efficiency. However, existing GC methods are exclusively designed for static graphs and lack the capability to process dynamic graphs with temporal dependencies.

\textbf{Dynamic Graph Neural Network} Recent research has extended conventional GNNs to dynamic graph neural networks (DGNNs)~\cite{DGNNsurvey, dgsurvey} by incorporating temporal modeling.  Owing to RNNs' powerful sequential processing capabilities, a dominant approach combines spatial GNNs with temporal RNN modules~\cite{LSTM} to capture evolving graph sequences~\cite{GC-LSTM, EvolveGCN, T-GCN, GCRN, STGCN}. In parallel, attention mechanisms~\cite{attention} have emerged as an effective alternative for modeling temporal dependencies across graph snapshots, offering adaptive weighting of temporal interactions~\cite{TGAT, TGN, DySAT}. Furthermore, various temporal modeling techniques~\cite{ROLAND} have been explored for DGNN architectures, including temporal point processes for event-based dynamics, spiking neural networks (SNN)~\cite{SNN} for bio-inspired temporal processing~\cite{dyrep}, and selective state space models (SSM)~\cite{GraphSSM} for efficient sequence modeling, etc. Although current DGNNs have demonstrated effectiveness in handling dynamic graph tasks, they inevitably incur non-negligible computational overhead when processing the time dimension.

\section{Proof of Proposition}\label{proof}

\textit{Proof of Proposition~\ref{pro}.}  
First, we prove sufficiency: if the snapshots are temporally independent, then the two expressions coincide. By the definition of KL divergence and its chain rule~\cite{CKL}, the divergence between the joint distributions of $\mathcal{T}$ and $\mathcal{S}$ can be expanded as:
\begin{equation}
\begin{aligned}
&\mathcal{D}_{\mathrm{KL}}\bigl(P_{G^{1:T}}\|\,Q_{\tilde G^{1:T}}\bigr)
= \sum_{g^{1:T}} p(g^{1:T}) \,\log\frac{p(g^{1:T})}{q(\tilde g^{1:T})} \\[4pt]
&= \sum_{g^{1:T}} p(g^{1:T})\,\log\frac{p(g^1)}{q(\tilde g^1)}
   \;+\; \sum_{g^{1:T}} p(g^{1:T}) \,\log\frac{\prod_{t=2}^T p(g^t\mid g^{1:t-1})}
                                            {\prod_{t=2}^T q(\tilde g^t\mid \tilde g^{1:t-1})} \\[4pt]
&= \sum_{g^1} p(g^1)\,\log\frac{p(g^1)}{q(\tilde g^1)}
   \;+\; \sum_{t=2}^T \sum_{g^{1:t-1}} p(g^{1:t-1})
      \Bigl[\sum_{g^t} p(g^t\mid g^{1:t-1}) 
             \log\frac{p(g^t\mid g^{1:t-1})}{q(\tilde g^t\mid \tilde g^{1:t-1})}\Bigr] \\[6pt]
&= \mathcal{D}_{\mathrm{KL}}\bigl(P_{G^1}\|\,Q_{\tilde G^1}\bigr)
   \;+\; \sum_{t=2}^T \mathcal{D}_{\mathrm{KL}}\bigl(P_{G^t\mid G^{1:t-1}}\|\,Q_{\tilde G^t\mid \tilde G^{1:t-1}}\bigr).
\end{aligned}
\end{equation}
Under the assumption of temporal independence, i.e., \( P_{G^{1:T}} = \prod_{t=1}^T P_{G^t} \) and \( Q_{\tilde{G}^{1:T}} = \prod_{t=1}^T Q_{\tilde{G}^t} \), each conditional distribution \( P_{G^t \mid G^{1:t-1}} \) reduces to the marginal \( P_{G^t} \), and likewise for \( Q \). As a result, the sum of conditional KL terms collapses  $\sum_{t=2}^T \mathcal{D}_{\mathrm{KL}}\bigl(P_{G^t}\,\|\,Q_{\tilde{G}^t}\bigr).$  Hence the joint KL over dynamic graph divergence collapses to:
\[
\mathcal{D}_{\mathrm{KL}}\bigl(P_{G^{1:T}}\|\,Q_{\tilde G^{1:T}}\bigr)
= \sum_{t=1}^T \mathcal{D}_{\mathrm{KL}}\bigl(P_{G^t}\|\,Q_{\tilde G^t}\bigr),
\]
which completes the proof of sufficiency.

Next, we construct a counterexample to demonstrate necessity. Specifically, we consider two dynamic graphs \(\mathcal{T}\) and \(\mathcal{S}\) over \(T = 2\) time steps, where the snapshots in \(\mathcal{T}\) are temporally independent, while those in \(\mathcal{S}\) exhibit strict temporal dependence. In detail:

\begin{itemize}[itemsep=1pt,leftmargin=20pt,topsep=1pt]
  \item The dynamic graph \(\mathcal{T} = \{G^t\}_{t=1}^2\) follows distribution \(P\), where each snapshot \(G^t \in \{g^0, g^1\}\) satisfies
    \[
      P(G^1 = g^0) = P(G^1 = g^1) = \tfrac12, 
      \quad
      P(G^2 = g^0) = P(G^2 = g^1) = \tfrac12,
    \]
    and \(G^1\) and \(G^2\) are independent, i.e.\ \(P(G^1, G^2) = P(G^1)\,P(G^2)\). Hence
    \[
      P(G^1 = g^i,\,G^2 = g^j) = \tfrac14,
      \quad
      \forall\,(g^i,g^j)\in\{g^0,g^1\}^2.
    \]

  \item The dynamic graph \(\mathcal{S} = \{\tilde G^t\}_{t=1}^2\) follows distribution \(Q\), where each snapshot \(\tilde G^t \in \{g^0, g^1\}\) has the same marginals
    \[
      Q(\tilde G^1 = g^0) = Q(\tilde G^1 = g^1) = \tfrac12,
      \quad
      Q(\tilde G^2 = g^0) = Q(\tilde G^2 = g^1) = \tfrac12,
    \]
    but \(\tilde G^2\) is perfectly dependent on \(\tilde G^1\) via \(\tilde G^2 = \tilde G^1\). Thus
    \[
      Q(\tilde G^1 = g^0,\,\tilde G^2 = g^0)
      = Q(\tilde G^1 = g^1,\,\tilde G^2 = g^1)
      = \tfrac12,
    \]
    and \(Q(\tilde G^1 = g^i,\,\tilde G^2 = g^j)=0\) for \(i\neq j\).
\end{itemize}

Observe that for both \(\mathcal{T}\) and \(\mathcal{S}\),
\[
\mathcal{D}_{\mathrm{KL}}\bigl(P_{G^t}\|\,Q_{\tilde G^t}\bigr)
= \tfrac12\log\frac{\tfrac12}{\tfrac12} + \tfrac12\log\frac{\tfrac12}{\tfrac12} = 0,
\quad t=1,2,
\]
so the sum of marginal divergences vanishes $\sum_{t=1}^2 \mathcal{D}_{\mathrm{KL}}(P_{G^t}\|\,Q_{\tilde G^t}) = 0.$ However, the joint KL divergence is:
\[
\mathcal{D}_{\mathrm{KL}}\bigl(P_{G^{1:2}}\,\|\,Q_{\tilde G^{1:2}}\bigr)
= \sum_{i,j\in\{0,1\}} \tfrac14 \log \frac{\tfrac14}{Q(\tilde G^1 = g^i, \tilde G^2 = g^j)}
= +\infty,
\]
This establishes that marginal alignment alone does not imply joint alignment unless temporal independence holds. 
\section{Algorithm Descriptions} \label{algorithm}
To provide a clearer understanding of how DyGC jointly optimizes the synthetic node features and evolving graph structures, we formally describe the entire condensation pipeline in Algorithm~\ref{alg:1}.  It jointly optimizes the synthetic features and the spiking structure generator via gradient-based updates, guided by both distribution and logit alignment losses.

\begin{algorithm}[ht]
\caption{Dynamic Graph Condensation (DyGC)}
\label{alg:1}
\begin{algorithmic}[1]
\REQUIRE 
\begin{itemize}
    \item[\textbf{(1)}] Real dynamic graph $\mathcal{T} = (\mathcal{A},\mathcal{X},\mathcal{Y})$ 
    \item[\textbf{(2)}] Synthetic node feature sequence $\tilde{\mathcal{X}}$ and pre-defined labels $\tilde{\mathcal{Y}}$;
    \item[\textbf{(3)}] Spiking structure generator $\delta_{\theta}$ with weight $\theta$;
    \item[\textbf{(4)}] A pretrained $\text{DGNN}_{\mathcal{T}}$ trained on $\mathcal{T}$;
    \item[\textbf{(6)}] Number of condensation loops $L$.
\end{itemize}
\ENSURE Condensed dynamic graph $\mathcal{S}^*$.
\FOR{$l = 0$ to $L-1$}
    \STATE Generate evolving structure for $\mathcal{S}$ by $\tilde{\mathcal{A}} = \delta_{\theta}(\tilde{\mathcal{X}})$
    \STATE Construct state evolving fields $\mathcal{H}^{\mathcal{T}}$ and $\mathcal{H}^{\mathcal{S}}$
    \STATE Compute $\mathcal{L}_{\text{dist}} =\sum_{c=1}^{C} \eta_{c} \left( 
 \mathcal{K}_c^{({\mathcal{T}}, {\mathcal{T}})}
- 2\mathcal{K}_c^{({\mathcal{T}}, {\mathcal{S}})}
+ \mathcal{K}_c^{({\mathcal{S}}, {\mathcal{S}})}
\right)$ \quad \quad \quad \quad 
    \STATE Generate soft labels  by $\tilde{\mathcal{Y}}' = \text{DGNN}_{\mathcal{T}}(\tilde{\mathcal{A}}, \tilde{\mathcal{X}})$
    \STATE  Compute $\mathcal{L}_{\text{logit}} =\sum_{i=1}^{m}-\tilde{\mathcal{Y}}_i\log \tilde{\mathcal{Y}}'_i$
    \STATE Update $\tilde{\mathcal{X}} \leftarrow \tilde{\mathcal{X}} - \eta_1\nabla_{\tilde{\mathcal{X}}}(\mathcal{L}_{\text{dist}} + \lambda \mathcal{L}_{\text{logit}})$
    \STATE Update $\mathcal{\theta} \leftarrow \mathcal{\theta} - \eta_2\nabla_{\mathcal{\theta}}(\mathcal{L}_{\text{dist}} + \lambda \mathcal{L}_{\text{logit}})$
\ENDFOR
\STATE $\tilde{\mathcal{A}} = \delta_{\theta}(\tilde{\mathcal{X}})$
\STATE \textbf{Return} $(\tilde{\mathcal{A}}, \tilde{\mathcal{X}}, \tilde{\mathcal{Y}})$
\end{algorithmic}
\end{algorithm}

It is worth noting that training the spiking structure generator presents a significant challenge due to the non-differentiability of discrete spikes and the hard threshold function.  To circumvent this issue, we follow the surrogate gradient method~\cite{PLIF}, which approximates the non-differentiable activation function with a smooth surrogate during backpropagation. In particular, we adopt the Sigmoid function $\sigma(\beta x) = 1 / (1 + \exp(-\beta x))$ as the surrogate, with its derivative defined as:
\begin{equation}
    \sigma'(\beta x) = \beta \cdot \sigma(\beta x) \cdot (1 - \sigma(\beta x)),
\end{equation}
where $\beta$ is a scaling factor controlling the sharpness of the approximation. While a larger $\beta$ yields a better approximation to the ideal hard threshold function $\Theta(\cdot)$, it may also lead to vanishing or exploding gradients, thereby hindering convergence. Hence, choosing a proper value of $\beta$ is critical for stable and effective training.

\section{Experimental Setup}\label{Experimental}
\subsection{Datasets.} The experiments are conducted on four dynamic graph benchmarks with different scales and time snapshots, including DBLP~\cite{SpikeNet}, Reddit~\cite{SpikeNet}, Arxiv~\cite{SpikeNet}, and Tmall~\cite{SpikeNet}. Dataset statistics are summarized in table~\ref{dataset} including the corresponding temporal continuity. The graph datasets are collected from real-world networks belonging to different domains.  DBLP and Arxiv are academic co-author networks extracted from  bibliography websites, in which
each node is an author and each edge means the two authors
collaborated on a paper. The authors in DBLP and Arxiv are labeled
according to their research areas.  Reddit is constructed from user-community interactions on the social platform Reddit, where each node represents either a user or a subreddit, and each edge indicates a user's participation through post submissions or comments.  Tmall is a bipartite graph extracted from the sales data at e-commerce platform Tmall, in which each node refers to either one user or one item and each edge refers to one purchase. The five most frequently purchased categories are treated as labels in experiments. Note that Tmall is a partially labeled dataset in which only items are assigned with categories.

\begin{table}[ht]
\centering
\caption{Dataset Statistics.}
\label{dataset}
\small
\begin{tabular}{l|cccc}
\toprule
\textbf{} & \textbf{DBLP}  & \textbf{Reddit} & \textbf{Arxiv} & \textbf{Tmall}  \\
\midrule
\textbf{\#Nodes} & 6,606  & 8,291 & 169,343 & 577,314 \\
\textbf{\#Edges} & 42,815 & 264,050 & 2,315,598  & 4,807,545\\
\textbf{\#Features} & 100 & 20 & 128 & 128 \\
\textbf{\#Classes} & 5 & 4 & 40 & 6 \\
\textbf{\#Time Steps} & 10 & 10 & 6 & 8 \\
\textbf{Category} & Citation & Society & Citation & E-commerce  \\
\bottomrule
\end{tabular}
\end{table}

\subsection{Backbone Architectures for DGNNs.} To evaluate the generalization capability of DyGC, we test the condensed dynamic graph on five representative DGNN architectures. These backbones span diverse design choices, combining static graph encoders with various sequence modeling strategies. The details are as follows:
\begin{itemize}[itemsep=1pt,leftmargin=30pt,topsep=1pt]
\item \textbf{T-GCN}~\cite{T-GCN} adopts a stacked architecture that combines GCN~\cite{GCN} and Gated Recurrent Unit (GRU), where GCN extract spatial features at each time step, and GRU model temporal dependencies across time steps.
\item \textbf{GCRN}~\cite{GCRN} integrates Long Short-Term Memory (LSTM) \cite{LSTM} units with parameterless graph convolution operations, enabling simultaneous modeling of spatial and temporal dependencies within a unified framework.
\item \textbf{STGCN}~\cite{STGCN} employs a spatiotemporal convolutional architecture that combines gated dilated causal convolutions for temporal modeling with graph convolutions for spatial feature learning.
\item \textbf{DySAT}~\cite{DySAT} uses a fully self-attentive architecture with attention mechanisms in both spatial and temporal dimensions, applying Graph Attention Networks (GATs)~\cite{GAT} for spatial encoding and Transformers~\cite{attention} for capturing temporal dependencies.
\item \textbf{ROLAND}~\cite{ROLAND} enhances the GNN-RNN framework by introducing hierarchical node states. It interleaves multiple GNN layers with temporal models, treating the embeddings from different GNN layers as hierarchical representations that evolve over time.
\end{itemize}

\subsection{Implementation details.} The DBLP, Reddit, and Tmall datasets are partitioned into training, validation, and testing sets with a 0.5/0.2/0.3 ratio. For the Arxiv dataset, we follow the official split provided by OGB. We evaluate model performance on the temporal node classification task using Macro-F1 and Micro-F1 scores. Macro-F1 calculates the unweighted mean of F1 scores across all classes, which is particularly useful for handling class imbalance. In contrast, Micro-F1 aggregates the contributions of all classes to provide an overall performance measure. Each experiment is repeated 5 times, and we report the mean and standard deviation (std) of the results to ensure robustness.

During the condensation process, node labels in the condensed dynamic graphs are assigned based on the class distribution of the original datasets and the specified condensation ratio. Node features are randomly initialized as learnable parameters. The learning rate for both the node features and the structure generator is uniformly set to 0.05 across all datasets. The temporal propagation coefficient $\alpha$ is fixed at 0.5 to establish a balanced temporal dependency. For spatiotemporal state matching, we consistently adopt the Gaussian RBF kernel to compute the MMD metric. In all experiments, we select T-GCN as the pretrained DGNN architecture for task-oriented alignment optimization.

For DGNN configurations, we set the hidden dimension to 128 for DBLP and Reddit, and 256 for Arxiv and Tmall. The number of graph encoder layers in DGNN is configured as 2, 1, 3, and 2 for the DBLP, Reddit, Arxiv, and Tmall datasets, respectively. All models, including our proposed method, baselines, and various DGNN architectures, are implemented using PyTorch and PyTorch Geometric, which are open-source frameworks released under the BSD and MIT licenses, respectively. All datasets used in the experiments are publicly available. Experiments are conducted on an NVIDIA RTX 4090 GPU with 24 GB of memory. The code will be included in the supplementary materials.

\subsection{Impact of Spatial Propagation Steps.}

\begin{figure}[htbp]
    \centering
    \includegraphics[width=0.8\linewidth]{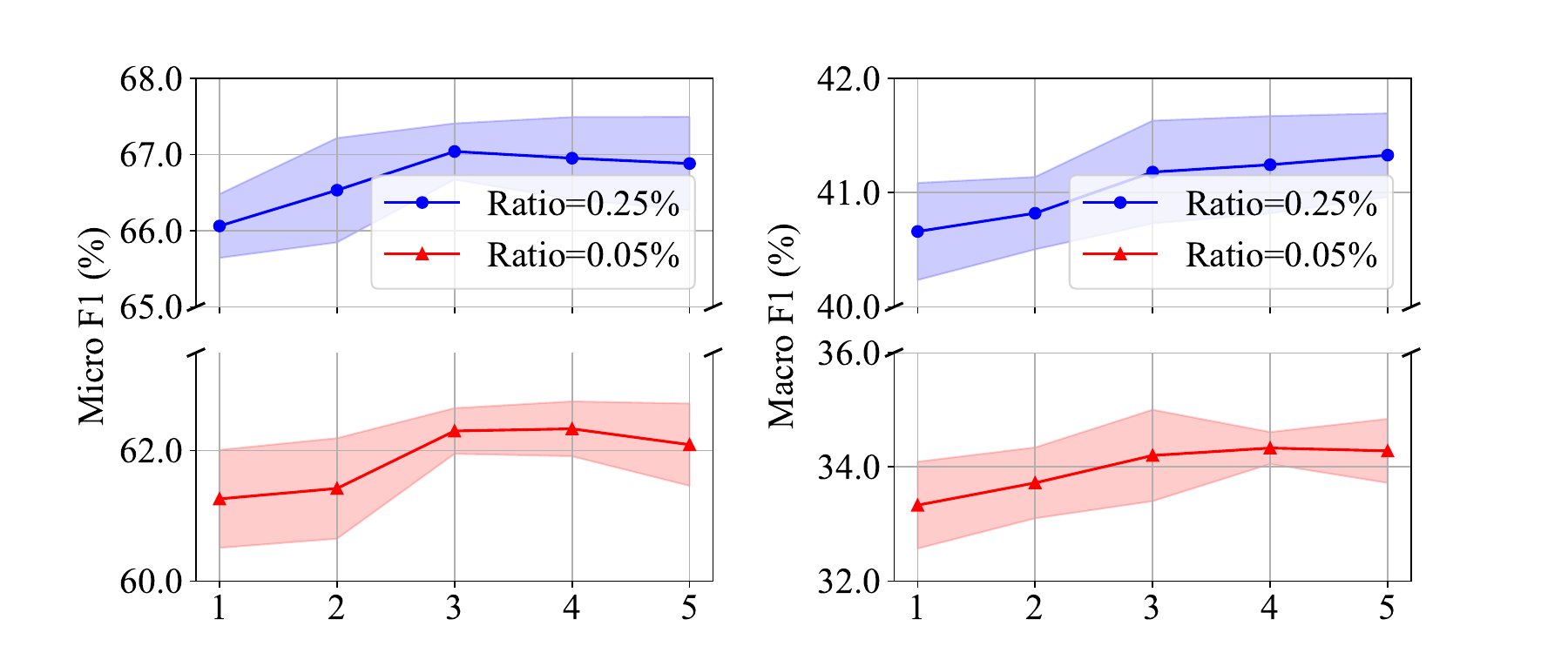}
    \caption{Micro-F1 (\%) and Macro-F1 (\%) results on the Arxiv with various spatial propagation steps $K$ and condensation ratios.}
    \label{order}
\end{figure}

We investigate the impact of different numbers of spatial propagation steps $K$ (ranging from 1 to 5) when constructing the state evolution field. As shown in Figure~\ref{order}, the performance improves noticeably when increasing $K$ from 1 to 3. This is because more propagation steps allow for introducing richer structural context into the state field, thereby achieving a more precise alignment of the distributions.  When $K$ exceeds 3, the performance tends to stabilize, exhibiting only minor fluctuations with slight improvements or decreases. This effect suggests that after sufficient neighborhood information has been integrated, additional propagation steps yield diminishing returns and may even cause information from preceding steps to be diluted during condensation.

\section{Discussions}\label{sec: discussions}
In this section, we discuss the limitations of the DyGC framework and propose a few future research directions that might be of interest.

\textbf{Discrete-time Assumption.}
At present, DyGC is designed for dynamic graphs represented as a sequence of discrete graph snapshots, where each snapshot corresponds to a specific and evenly spaced time step. This setting simplifies modeling and training, and aligns with existing DGNN architectures. However, many real-world graphs such as financial transaction networks, communication logs, and user interaction graphs evolve in continuous time, where events occur asynchronously and irregularly. In such settings, the discretization process may result in loss of temporal precision. Addressing this limitation requires rethinking the core components of DyGC and condensation objectives. One promising direction is to incorporate temporal point process  or continuous-time message passing frameworks to enable condensation in high-resolution, event-based dynamic graphs.

\textbf{Task Specificity to Classification.}  
The current DyGC framework is primarily optimized for temporal node classification tasks, where the goal is to preserve discriminative node-level information across time. While node classification is a foundational benchmark in dynamic graph learning, it does not capture the full range of temporal graph analytics. In practice, applications such as dynamic link prediction, temporal community detection, time-aware recommendation, and dynamic graph regression are of significant interest and often involve different inductive biases. Extending DyGC to support such tasks may require re-designing the condensation criteria, supervision signals, and distribution matching strategies. For instance, preserving edge evolution patterns or long-range temporal dependencies may be more critical in link prediction tasks than in node classification.



\newpage
\section*{NeurIPS Paper Checklist}

\begin{enumerate}

\item {\bf Claims}
    \item[] Question: Do the main claims made in the abstract and introduction accurately reflect the paper's contributions and scope?
    \item[] Answer: \answerYes{} 
    \item[] Justification: The main claims made in the abstract and introduction has accurately reflected the paper's contributions and scope.
    \item[] Guidelines:
    \begin{itemize}
        \item The answer NA means that the abstract and introduction do not include the claims made in the paper.
        \item The abstract and/or introduction should clearly state the claims made, including the contributions made in the paper and important assumptions and limitations. A No or NA answer to this question will not be perceived well by the reviewers. 
        \item The claims made should match theoretical and experimental results, and reflect how much the results can be expected to generalize to other settings. 
        \item It is fine to include aspirational goals as motivation as long as it is clear that these goals are not attained by the paper. 
    \end{itemize}

\item {\bf Limitations}
    \item[] Question: Does the paper discuss the limitations of the work performed by the authors?
    \item[] Answer: \answerYes{} 
    \item[] Justification: We have discussed the limitation of this work in appendix \ref{sec: discussions}.
    \item[] Guidelines:
    \begin{itemize}
        \item The answer NA means that the paper has no limitation while the answer No means that the paper has limitations, but those are not discussed in the paper. 
        \item The authors are encouraged to create a separate "Limitations" section in their paper.
        \item The paper should point out any strong assumptions and how robust the results are to violations of these assumptions (e.g., independence assumptions, noiseless settings, model well-specification, asymptotic approximations only holding locally). The authors should reflect on how these assumptions might be violated in practice and what the implications would be.
        \item The authors should reflect on the scope of the claims made, e.g., if the approach was only tested on a few datasets or with a few runs. In general, empirical results often depend on implicit assumptions, which should be articulated.
        \item The authors should reflect on the factors that influence the performance of the approach. For example, a facial recognition algorithm may perform poorly when image resolution is low or images are taken in low lighting. Or a speech-to-text system might not be used reliably to provide closed captions for online lectures because it fails to handle technical jargon.
        \item The authors should discuss the computational efficiency of the proposed algorithms and how they scale with dataset size.
        \item If applicable, the authors should discuss possible limitations of their approach to address problems of privacy and fairness.
        \item While the authors might fear that complete honesty about limitations might be used by reviewers as grounds for rejection, a worse outcome might be that reviewers discover limitations that aren't acknowledged in the paper. The authors should use their best judgment and recognize that individual actions in favor of transparency play an important role in developing norms that preserve the integrity of the community. Reviewers will be specifically instructed to not penalize honesty concerning limitations.
    \end{itemize}

\item {\bf Theory Assumptions and Proofs}
    \item[] Question: For each theoretical result, does the paper provide the full set of assumptions and a complete (and correct) proof?
    \item[] Answer: \answerYes{} 
    \item[] Justification: All the theorems, formulas, and proofs in the paper are numbered and cross-referenced. The proof of theorems are presented in appendix\ref{proof}.
    \item[] Guidelines:
    \begin{itemize}
        \item The answer NA means that the paper does not include theoretical results. 
        \item All the theorems, formulas, and proofs in the paper should be numbered and cross-referenced.
        \item All assumptions should be clearly stated or referenced in the statement of any theorems.
        \item The proofs can either appear in the main paper or the supplemental material, but if they appear in the supplemental material, the authors are encouraged to provide a short proof sketch to provide intuition. 
        \item Inversely, any informal proof provided in the core of the paper should be complemented by formal proofs provided in appendix or supplemental material.
        \item Theorems and Lemmas that the proof relies upon should be properly referenced. 
    \end{itemize}

    \item {\bf Experimental Result Reproducibility}
    \item[] Question: Does the paper fully disclose all the information needed to reproduce the main experimental results of the paper to the extent that it affects the main claims and/or conclusions of the paper (regardless of whether the code and data are provided or not)?
    \item[] Answer: \answerYes{} 
    \item[] Justification: We provide a comprehensive description of the experimental settings in appendix \ref{Experimental}. The algorithm descriptions of DyGC is presented in appendix \ref{algorithm}.
    All the code for reproducing the experiments is made available in the supplementary material accompanying the submission.
    \item[] Guidelines:
    \begin{itemize}
        \item The answer NA means that the paper does not include experiments.
        \item If the paper includes experiments, a No answer to this question will not be perceived well by the reviewers: Making the paper reproducible is important, regardless of whether the code and data are provided or not.
        \item If the contribution is a dataset and/or model, the authors should describe the steps taken to make their results reproducible or verifiable. 
        \item Depending on the contribution, reproducibility can be accomplished in various ways. For example, if the contribution is a novel architecture, describing the architecture fully might suffice, or if the contribution is a specific model and empirical evaluation, it may be necessary to either make it possible for others to replicate the model with the same dataset, or provide access to the model. In general. releasing code and data is often one good way to accomplish this, but reproducibility can also be provided via detailed instructions for how to replicate the results, access to a hosted model (e.g., in the case of a large language model), releasing of a model checkpoint, or other means that are appropriate to the research performed.
        \item While NeurIPS does not require releasing code, the conference does require all submissions to provide some reasonable avenue for reproducibility, which may depend on the nature of the contribution. For example
        \begin{enumerate}
            \item If the contribution is primarily a new algorithm, the paper should make it clear how to reproduce that algorithm.
            \item If the contribution is primarily a new model architecture, the paper should describe the architecture clearly and fully.
            \item If the contribution is a new model (e.g., a large language model), then there should either be a way to access this model for reproducing the results or a way to reproduce the model (e.g., with an open-source dataset or instructions for how to construct the dataset).
            \item We recognize that reproducibility may be tricky in some cases, in which case authors are welcome to describe the particular way they provide for reproducibility. In the case of closed-source models, it may be that access to the model is limited in some way (e.g., to registered users), but it should be possible for other researchers to have some path to reproducing or verifying the results.
        \end{enumerate}
    \end{itemize}

\item {\bf Open access to data and code}
    \item[] Question: Does the paper provide open access to the data and code, with sufficient instructions to faithfully reproduce the main experimental results, as described in supplemental material?
    \item[] Answer: \answerYes{} 
    \item[] Justification: All the data used in our experiments are publicly available online and the code to reproduce the experiments is available in the supplementary material accompanying the submission.
    \item[] Guidelines:
    \begin{itemize}
        \item The answer NA means that paper does not include experiments requiring code.
        \item Please see the NeurIPS code and data submission guidelines (\url{https://nips.cc/public/guides/CodeSubmissionPolicy}) for more details.
        \item While we encourage the release of code and data, we understand that this might not be possible, so “No” is an acceptable answer. Papers cannot be rejected simply for not including code, unless this is central to the contribution (e.g., for a new open-source benchmark).
        \item The instructions should contain the exact command and environment needed to run to reproduce the results. See the NeurIPS code and data submission guidelines (\url{https://nips.cc/public/guides/CodeSubmissionPolicy}) for more details.
        \item The authors should provide instructions on data access and preparation, including how to access the raw data, preprocessed data, intermediate data, and generated data, etc.
        \item The authors should provide scripts to reproduce all experimental results for the new proposed method and baselines. If only a subset of experiments are reproducible, they should state which ones are omitted from the script and why.
        \item At submission time, to preserve anonymity, the authors should release anonymized versions (if applicable).
        \item Providing as much information as possible in supplemental material (appended to the paper) is recommended, but including URLs to data and code is permitted.
    \end{itemize}

\item {\bf Experimental Setting/Details}
    \item[] Question: Does the paper specify all the training and test details (e.g., data splits, hyperparameters, how they were chosen, type of optimizer, etc.) necessary to understand the results?
    \item[] Answer: \answerYes{} 
    \item[] Justification: We have provided a comprehensive description of the experimental settings in appendix \ref{Experimental}.
    \item[] Guidelines:
    \begin{itemize}
        \item The answer NA means that the paper does not include experiments.
        \item The experimental setting should be presented in the core of the paper to a level of detail that is necessary to appreciate the results and make sense of them.
        \item The full details can be provided either with the code, in appendix, or as supplemental material.
    \end{itemize}

\item {\bf Experiment Statistical Significance}
    \item[] Question: Does the paper report error bars suitably and correctly defined or other appropriate information about the statistical significance of the experiments?
    \item[] Answer: \answerYes{} 
    \item[] Justification: The experiments were conducted over 5 runs, and we present the averaged results along with the standard deviation.
    \item[] Guidelines:
    \begin{itemize}
        \item The answer NA means that the paper does not include experiments.
        \item The authors should answer "Yes" if the results are accompanied by error bars, confidence intervals, or statistical significance tests, at least for the experiments that support the main claims of the paper.
        \item The factors of variability that the error bars are capturing should be clearly stated (for example, train/test split, initialization, random drawing of some parameter, or overall run with given experimental conditions).
        \item The method for calculating the error bars should be explained (closed form formula, call to a library function, bootstrap, etc.)
        \item The assumptions made should be given (e.g., Normally distributed errors).
        \item It should be clear whether the error bar is the standard deviation or the standard error of the mean.
        \item It is OK to report 1-sigma error bars, but one should state it. The authors should preferably report a 2-sigma error bar than state that they have a 96\% CI, if the hypothesis of Normality of errors is not verified.
        \item For asymmetric distributions, the authors should be careful not to show in tables or figures symmetric error bars that would yield results that are out of range (e.g. negative error rates).
        \item If error bars are reported in tables or plots, The authors should explain in the text how they were calculated and reference the corresponding figures or tables in the text.
    \end{itemize}

\item {\bf Experiments Compute Resources}
    \item[] Question: For each experiment, does the paper provide sufficient information on the computer resources (type of compute workers, memory, time of execution) needed to reproduce the experiments?
    \item[] Answer: \answerYes{} 
    \item[] Justification: Implementation details including software and hardware infrastructures are listed in appendix \ref{Experimental}.
    \item[] Guidelines:
    \begin{itemize}
        \item The answer NA means that the paper does not include experiments.
        \item The paper should indicate the type of compute workers CPU or GPU, internal cluster, or cloud provider, including relevant memory and storage.
        \item The paper should provide the amount of compute required for each of the individual experimental runs as well as estimate the total compute. 
        \item The paper should disclose whether the full research project required more compute than the experiments reported in the paper (e.g., preliminary or failed experiments that didn't make it into the paper). 
    \end{itemize}
    
\item {\bf Code Of Ethics}
    \item[] Question: Does the research conducted in the paper conform, in every respect, with the NeurIPS Code of Ethics \url{https://neurips.cc/public/EthicsGuidelines}?
    \item[] Answer: \answerYes{} 
    \item[] Justification: The attached code has undergone thorough scrutiny to guarantee anonymity and adherence to the NeurIPS Code of Ethics.
    \item[] Guidelines:
    \begin{itemize}
        \item The answer NA means that the authors have not reviewed the NeurIPS Code of Ethics.
        \item If the authors answer No, they should explain the special circumstances that require a deviation from the Code of Ethics.
        \item The authors should make sure to preserve anonymity (e.g., if there is a special consideration due to laws or regulations in their jurisdiction).
    \end{itemize}

\item {\bf Broader Impacts}
    \item[] Question: Does the paper discuss both potential positive societal impacts and negative societal impacts of the work performed?
    \item[] Answer: \answerYes{} 
    \item[] Justification: The discussion on both potential positive societal impacts and negative societal impacts of the work is provided in appendix \ref{appendix:broader_impact}.
    \item[] Guidelines:
    \begin{itemize}
        \item The answer NA means that there is no societal impact of the work performed.
        \item If the authors answer NA or No, they should explain why their work has no societal impact or why the paper does not address societal impact.
        \item Examples of negative societal impacts include potential malicious or unintended uses (e.g., disinformation, generating fake profiles, surveillance), fairness considerations (e.g., deployment of technologies that could make decisions that unfairly impact specific groups), privacy considerations, and security considerations.
        \item The conference expects that many papers will be foundational research and not tied to particular applications, let alone deployments. However, if there is a direct path to any negative applications, the authors should point it out. For example, it is legitimate to point out that an improvement in the quality of generative models could be used to generate deepfakes for disinformation. On the other hand, it is not needed to point out that a generic algorithm for optimizing neural networks could enable people to train models that generate Deepfakes faster.
        \item The authors should consider possible harms that could arise when the technology is being used as intended and functioning correctly, harms that could arise when the technology is being used as intended but gives incorrect results, and harms following from (intentional or unintentional) misuse of the technology.
        \item If there are negative societal impacts, the authors could also discuss possible mitigation strategies (e.g., gated release of models, providing defenses in addition to attacks, mechanisms for monitoring misuse, mechanisms to monitor how a system learns from feedback over time, improving the efficiency and accessibility of ML).
    \end{itemize}
    
\item {\bf Safeguards}
    \item[] Question: Does the paper describe safeguards that have been put in place for responsible release of data or models that have a high risk for misuse (e.g., pretrained language models, image generators, or scraped datasets)?
    \item[] Answer: \answerNA{} 
    \item[] Justification: This paper poses no such risks.
    \item[] Guidelines:
    \begin{itemize}
        \item The answer NA means that the paper poses no such risks.
        \item Released models that have a high risk for misuse or dual-use should be released with necessary safeguards to allow for controlled use of the model, for example by requiring that users adhere to usage guidelines or restrictions to access the model or implementing safety filters. 
        \item Datasets that have been scraped from the Internet could pose safety risks. The authors should describe how they avoided releasing unsafe images.
        \item We recognize that providing effective safeguards is challenging, and many papers do not require this, but we encourage authors to take this into account and make a best faith effort.
    \end{itemize}

\item {\bf Licenses for existing assets}
    \item[] Question: Are the creators or original owners of assets (e.g., code, data, models), used in the paper, properly credited and are the license and terms of use explicitly mentioned and properly respected?
    \item[] Answer: \answerYes{} 
    \item[] Justification: We have cited the original paper that produced the code package or dataset and have explicitly stated the license used for the open-source frameworks.
    \item[] Guidelines:
    \begin{itemize}
        \item The answer NA means that the paper does not use existing assets.
        \item The authors should cite the original paper that produced the code package or dataset.
        \item The authors should state which version of the asset is used and, if possible, include a URL.
        \item The name of the license (e.g., CC-BY 4.0) should be included for each asset.
        \item For scraped data from a particular source (e.g., website), the copyright and terms of service of that source should be provided.
        \item If assets are released, the license, copyright information, and terms of use in the package should be provided. For popular datasets, \url{paperswithcode.com/datasets} has curated licenses for some datasets. Their licensing guide can help determine the license of a dataset.
        \item For existing datasets that are re-packaged, both the original license and the license of the derived asset (if it has changed) should be provided.
        \item If this information is not available online, the authors are encouraged to reach out to the asset's creators.
    \end{itemize}

\item {\bf New Assets}
    \item[] Question: Are new assets introduced in the paper well documented and is the documentation provided alongside the assets?
    \item[] Answer: \answerNA{} 
    \item[] Justification: This paper does not release new assets.
    \item[] Guidelines:
    \begin{itemize}
        \item The answer NA means that the paper does not release new assets.
        \item Researchers should communicate the details of the dataset/code/model as part of their submissions via structured templates. This includes details about training, license, limitations, etc. 
        \item The paper should discuss whether and how consent was obtained from people whose asset is used.
        \item At submission time, remember to anonymize your assets (if applicable). You can either create an anonymized URL or include an anonymized zip file.
    \end{itemize}

\item {\bf Crowdsourcing and Research with Human Subjects}
    \item[] Question: For crowdsourcing experiments and research with human subjects, does the paper include the full text of instructions given to participants and screenshots, if applicable, as well as details about compensation (if any)? 
    \item[] Answer: \answerNA{} 
    \item[] Justification: This paper does not involve crowdsourcing nor research with human subjects.
    \item[] Guidelines:
    \begin{itemize}
        \item The answer NA means that the paper does not involve crowdsourcing nor research with human subjects.
        \item Including this information in the supplemental material is fine, but if the main contribution of the paper involves human subjects, then as much detail as possible should be included in the main paper. 
        \item According to the NeurIPS Code of Ethics, workers involved in data collection, curation, or other labor should be paid at least the minimum wage in the country of the data collector. 
    \end{itemize}

\item {\bf Institutional Review Board (IRB) Approvals or Equivalent for Research with Human Subjects}
    \item[] Question: Does the paper describe potential risks incurred by study participants, whether such risks were disclosed to the subjects, and whether Institutional Review Board (IRB) approvals (or an equivalent approval/review based on the requirements of your country or institution) were obtained?
    \item[] Answer: \answerNA{} 
    \item[] Justification: This paper does not involve crowdsourcing nor research with human subjects.
    \item[] Guidelines:
    \begin{itemize}
        \item The answer NA means that the paper does not involve crowdsourcing nor research with human subjects.
        \item Depending on the country in which research is conducted, IRB approval (or equivalent) may be required for any human subjects research. If you obtained IRB approval, you should clearly state this in the paper. 
        \item We recognize that the procedures for this may vary significantly between institutions and locations, and we expect authors to adhere to the NeurIPS Code of Ethics and the guidelines for their institution. 
        \item For initial submissions, do not include any information that would break anonymity (if applicable), such as the institution conducting the review.
    \end{itemize}

\end{enumerate}

\end{document}